\documentclass[10pt,twocolumn,letterpaper]{m3ailpreprint}

\usepackage{xspace}

\makeatletter
\DeclareRobustCommand\onedot{\futurelet\@let@token\@onedot}
\def\@onedot{\ifx\@let@token.\else.\null\fi\xspace}
\def\eg{\emph{e.g}\onedot} 
\makeatother

\colorlet{tbhead}{mthreeblue!8}
\colorlet{tbzebra}{mthreeblue!3}
\newcommand{\headrulefill}{\noalign{{\color{tbhead}\hrule height\cmidrulewidth}\vskip-\cmidrulewidth}}
\newcommand{\meanstd}[2]{\ensuremath{#1_{\mbox{\scriptsize$\pm$ #2}}}}
\newcommand{\secondbest}[1]{\mbox{\setbox0\hbox{#1}\rlap{\rule[-1.6pt]{\wd0}{0.7pt}}\box0}}

\begin{document}
\title{What Truly Grounds a Concept? Evidence-Aligned Concept Bottleneck Model}

\author[1]{Di Yan}
\author[2]{Beibei Feng}
\author[2]{Yan Yang}
\author[1,\dagger]{Qiang Huang}
\author[1]{Jun Yu}

\affiliation[1]{Harbin Institute of Technology (Shenzhen)}
\affiliation[2]{Hangzhou Dianzi University}

\contribution{$^\dagger$Corresponding author.}

\checkdata[Correspondence]{\email{huangqiang@hit.edu.cn}}

\abstract{%
Concept Bottleneck Models (CBMs) route classifications through human-readable concepts, enabling inspection of classifier decisions.
Yet correct concept predictions can rely on co-occurring parts or context rather than the intended visual evidence.
We introduce \textbf{EA-CBM} (\textbf{E}vidence-\textbf{A}ligned \textbf{C}oncept \textbf{B}ottleneck \textbf{M}odel), which averages one spatial evidence map per concept into the bottleneck used for classification.
Region-filter matching and spatial concept grounding align these maps with annotated regions.
Matched regional perturbations encourage larger response drops under target-region removal than under control-region removal, while limiting control-induced changes.
Region annotations are used only during training; inference requires the image alone.
Across six datasets with concept-region annotations, EA-CBM achieves the highest concept mAP and Concept Trustworthiness Score (CTS) among twelve CBM baselines, improving CTS over each dataset's strongest method by 8.6--42.2 percentage points.
Classification accuracy remains competitive with concept-supervised CBMs.
With concept labels and masks on only 5\% of training images and class supervision elsewhere, EA-CBM surpasses every baseline in CTS on all six datasets with regions.
}

\maketitle

\section{Introduction}

Concept Bottleneck Models (CBMs) predict classes using named concepts, enabling users to inspect intermediate predictions~\cite{koh2020concept} and correct them via concept intervention~\cite{koh2020concept, zarlenga2023learning}.
CBMs have expanded from supervised attribute predictors~\cite{koh2020concept, zarlenga2022concept} to methods that acquire concepts through language models and vision--language similarity~\cite{oikarinen2023label, yang2023language, srivastava2024vlg} or expose spatial concept evidence~\cite{huang2024concept, zhang2026slcbm, colamonaco2026prototype}.

\begin{figure}[!t]
\centering
\includegraphics[width=1.0\columnwidth]{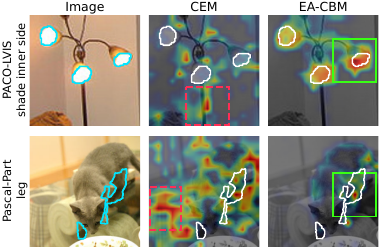}%
\vspace{-0.5em}
\caption{\textbf{Concept accuracy does not ensure spatial alignment.} CEM correctly predicts \emph{shade inner side} and \emph{leg} but peaks on the lamp stem and background; EA-CBM peaks within annotated regions (contours). Boxes: CTS windows (green: hit; red: miss).}
\label{fig:teaser}
\vspace{-1.0em}
\end{figure}

For these intermediate predictions to support meaningful visual inspection, each concept response should reflect the property it names.
Image-level concept labels supervise whether a property is present, but do not specify which visual evidence should support its prediction.
A model can therefore predict concepts from co-occurring parts or surrounding context~\cite{mahinpei2021promises, margeloiu2021concept, huang2024concept, raman2025localities}.
Figure~\ref{fig:teaser} illustrates this ambiguity: CEM correctly predicts a lamp's shade and a cat's leg, yet its attributions peak on the stem and background.
Concept accuracy alone thus does not establish evidence alignment.
Such responses can mislead inspection, and correcting a bottleneck value does not by itself change the evidence used by the learned concept predictor.

Region annotations provide information missing from image-level labels: where the named property appears.
Many concept datasets supply this information through object-part masks, face-part masks, or expert boxes~\cite{ramanathan2023paco, chen2014detect, lee2020maskgan, lamas2021monumai}.
Existing approaches use spatial information in different ways: spatial concept heads expose evidence locations, PGCM~\cite{colamonaco2026prototype} can supervise its part segmenter with masks, and explanation-supervision methods~\cite{ross2017right, gao2022res} regularize class-output attributions.
We connect region supervision directly to the concept responses used for classification.
We seek responses whose evidence aligns with the annotated region and whose value drops selectively when the evidence is removed.

Achieving these complementary goals with limited annotations presents three challenges:
\begin{itemize}
  \item \textbf{Learning from the Correct Regions.} 
  Co-occurring parts and context can trigger concept responses in wrong regions. Training must associate each named property with its own region and concentrate positive evidence there.
  
  \item \textbf{Learning Evidence-Dependent Responses.} 
  Spatial alignment alone does not establish regional dependence: local features may encode nonlocal information, and any removal can alter responses through general image disruption.
  Training must separate target-evidence removal from matched non-target perturbations.
  
  \item \textbf{Learning from Few Annotations.} 
  Region masks are costly, and concept labels and masks may cover only a small subset of training images. Models must learn evidence alignment from this limited supervision while retaining accuracy.
\end{itemize}

We introduce \textbf{EA-CBM} (\textbf{E}vidence-\textbf{A}ligned \textbf{C}oncept \textbf{B}ottleneck \textbf{M}odel), which directly links spatial evidence supervision to the concept bottleneck used for classification.
Each concept has an evidence map whose spatial average forms one bottleneck coordinate, providing an explicit spatial decomposition of the response supplied to the classifier.
Region-filter matching links region features to concept filters, while spatial concept grounding concentrates positive evidence in annotated regions.
Matched regional perturbations encourage selective response decreases when the corresponding evidence is removed.
These objectives use region annotations only in training; inference retains the image-to-concept-to-label path and needs only the image.

Across six datasets with concept-region annotations, EA-CBM achieves the highest Concept Trustworthiness Score (CTS) and concept mAP among twelve CBM baselines, improving CTS over each dataset's strongest baseline by 8.6--42.2 percentage points.
Classification remains competitive with concept-supervised CBMs across all eight datasets, where EA-CBM uses only concept and class losses with class-level concept lists.
Ablations link localization gains to region objectives, region-removal tests support improved response selectivity, and qualitative maps show concentrated evidence on small parts baselines miss (Figure~\ref{fig:case}).

Our contributions are summarized as follows:
\begin{itemize}
  \item \textbf{Region-Supervised Concept Learning.} 
  We directly supervise the filters and evidence maps whose pooled responses form the concept bottleneck.
  Region-filter matching links annotated region features to their concepts, while spatial concept grounding encourages positive evidence inside concept regions and suppresses it outside.
  
  \item \textbf{Matched Regional Perturbations.} 
  We train response selectivity by comparing target-region perturbations with non-target perturbations in the same image.
  The objective encourages larger concept-response decreases for target removal while limiting changes under controls, complementing spatial alignment with supervision of the pooled response.

  \item \textbf{Learning from Limited Annotations.} 
  EA-CBM uses available concept and region supervision, with class supervision throughout.
  With both annotations on only 5\% of training images, it exceeds the strongest baseline under its standard supervision by 3.4--29.3 CTS points across six datasets with regions, while accuracy remains within 1.6 points of fully annotated EA-CBM.
\end{itemize}

\section{Related Work}

\paragraph{Concept Bottleneck Models (CBMs)}
CBMs~\cite{koh2020concept} predict classes using human-readable concepts.
Extensions enrich bottlenecks with auxiliary concepts~\cite{sawada2022concept, sheth2023auxiliary, zhang2025decoupling}, concept activation vectors~\cite{yuksekgonul2023post}, or embeddings~\cite{zarlenga2022concept, zarlenga2023learning, zarlenga2025avoiding}, and model uncertainty and dependencies~\cite{kim2023probabilistic, vandenhirtz2024stochastic, hoffmann2025posthoc, xu2024energy, dominici2025counterfactual}.
Other work improves interventions~\cite{chauhan2023interactive}, reduces concept-label requirements~\cite{hu2025semi, selvaraj2024alignment}, and constrains bottleneck information flow~\cite{almudevar2026never, galliamov2026cibm}. 
However, image-level concept agreement alone does not establish whether responses rely on the intended visual evidence.

Label-free and concept-discovery methods reduce manual annotation by scoring textual concepts with vision-language similarity~\cite{yang2023language, yan2023learning, panousis2023sparse, he2025v2c, tan2024explain, jiang2025enhancing, zhang2025attribute}, learning image-to-concept projections from model-generated targets~\cite{oikarinen2023label, lai2024faithful, srivastava2024vlg, xie2026hierarchical, moayeri2023text, sammani2026clipfree, zhao2026partially, du2026flexible, semenov2024sparse, yu2025language, tapli2026rethinking}, decomposing visual features into textual concepts~\cite{bhalla2024interpreting, gong2026concepts, wang2025cbmzero}, or discovering concepts within visual representations~\cite{rao2024discover, schrodi2025selective, prasse2025dcbm, wang2023bottleneck, shang2024incremental, desantis2026mechanistic, liu2025hybrid}.
Beyond concept acquisition and representation, EA-CBM uses annotated regions to align the evidence maps of its bottleneck and matched regional perturbations to encourage selective dependence on those regions.

\begin{figure*}[t]
  \centering
  \includegraphics[width=1.0\textwidth]{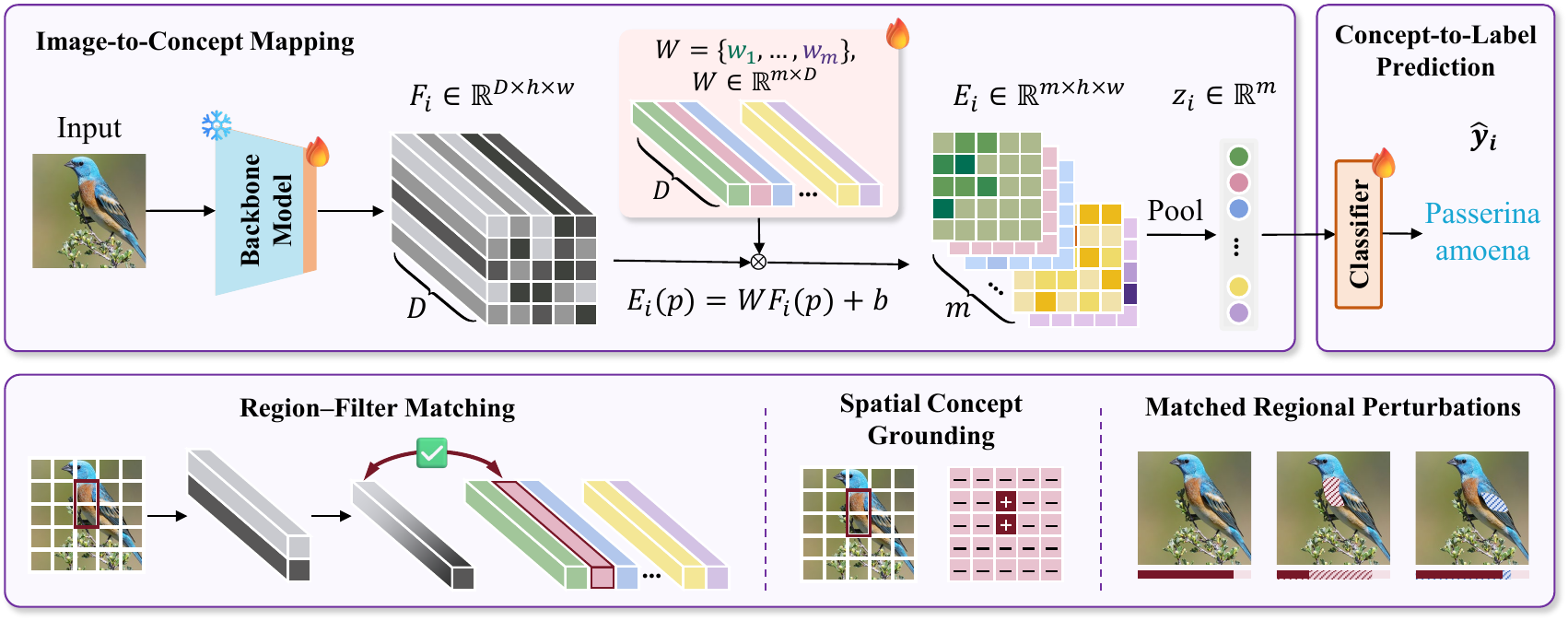}
  \vspace{-1.5em}
  \caption{\textbf{EA-CBM architecture and region objectives.}
  Top: concept filters map spatial visual features to evidence maps, which are pooled into a concept vector (Section~\ref{sec:eacbm:mapping}) for class prediction (Section~\ref{sec:eacbm:prediction}). Snowflakes and flames denote frozen and trainable components. The final encoder block, concept filters, and classifier are trained jointly with concept, class, and region objectives.
  Bottom, from left to right: the training-only objectives of region-filter matching, spatial concept grounding, and matched regional perturbations (Section~\ref{sec:eacbm:training_obj}).}
  \label{fig:eacbm_architecture}
  \vspace{-0.5em}
\end{figure*}

\paragraph{Spatial Evidence for Concepts}
Spatial CBMs expose the image regions associated with concepts.
Following ProtoPNet~\cite{chen2019looks}, ProtoCBM~\cite{huang2024concept} predicts concepts from local prototype matches, while PGCM~\cite{colamonaco2026prototype} assigns segmented image parts to prototypes with explicit concept representations.
SL-CBM~\cite{zhang2026slcbm} pools spatial concept maps, DOT-CBM~\cite{xie2025discovering} aligns patches with concepts through optimal transport, Coarse-to-Fine CBM~\cite{panousis2024coarse} and SALF-CBM~\cite{benou2025show} derive concepts from local image features, and OCB~\cite{steinmann2025object} organizes concepts around detected objects.

Spatial evidence alone does not establish regional dependence; studies of leakage~\cite{mahinpei2021promises, margeloiu2021concept, havasi2022addressing}, locality~\cite{raman2025localities, kumar2026measuring}, and faithfulness~\cite{schmalwasser2026faithfulness} identify complementary limitations of concept explanations.
We evaluate localization with ProtoCBM's Concept Trustworthiness Score~\cite{huang2024concept}, while EA-CBM trains both spatial alignment and selective response changes under regional perturbations.

\paragraph{Region and Explanation Supervision}
RRR~\cite{ross2017right} and RES~\cite{gao2022res} regularize class-output explanations by penalizing input gradients in annotated irrelevant regions and aligning attribution maps with imperfect human explanations, respectively.
Within CBMs, PGCM~\cite{colamonaco2026prototype} can train its part segmenter with masks, while VLG-CBM~\cite{srivastava2024vlg} derives concept targets and crop augmentations from detector boxes.
Other approaches ground concept groups in lesion-derived zones~\cite{tonmoy2026spatially}, guide part-specific attention with keypoint-derived priors~\cite{ramachandram2026spatially}, or aggregate concept evidence from automatically grounded segments~\cite{eisenberg2025spatially}.

EA-CBM combines region-filter matching, spatial grounding, and matched perturbations that encourage larger concept-response drops under target-region removal than under control-region removal, while limiting control-induced changes.
Using region annotations only during training, these objectives supervise the evidence maps and their spatial averages that form the classifier's bottleneck.
\section{EA-CBM}
\label{sec:eacbm}

Concept labels specify which properties an image shows, but not the visual evidence supporting them.
EA-CBM addresses this ambiguity by supervising both concept-evidence localization and responses to regional perturbations.
As shown in Figure~\ref{fig:eacbm_architecture}, each concept has a spatial evidence map whose average forms one coordinate of a vector-valued bottleneck.
Three complementary objectives guide learning: region--filter matching associates region features with their named concepts, spatial concept grounding concentrates evidence within annotated regions, and matched regional perturbations encourage selective dependence on those regions.
Region annotations are used only during training; inference requires the image alone.

\subsection{Image-to-Concept Mapping}
\label{sec:eacbm:mapping}

To make spatial evidence directly accessible to supervision, we retain spatial features through the concept layer and pool the resulting evidence maps into concept responses (Figure~\ref{fig:eacbm_architecture}, left).
For image \(x_i\), a pretrained visual encoder produces features \(\bm{F}_i\in\mathbb{R}^{D \times h \times w}\) on grid \(\Omega_F\), with a \(D\)-dimensional representation at each location.
We freeze the earlier encoder blocks and jointly update the final transformer block and concept layer, allowing the spatial features to adapt to concept and region supervision.
With CLIP ViT-L/14, \(\bm{F}_i\) comprises \(16\times16\) final-block patch tokens with \(D=1024\). 
For $m$ concepts, the image-to-concept mapping therefore trains one transformer block and \(m(D+1)\) concept-filter parameters.

Each concept $c$ consists of a learned filter \(\bm{w}_c\in\mathbb{R}^{D}\) and bias \(b_c\).
Stacked as \(\bm{W}\in\mathbb{R}^{m\times D}\), these filters form a shared \(1\times1\) convolution that produces evidence maps \(\bm{E}_i\in\mathbb{R}^{m\times h\times w}\).
Global average pooling converts each map into a concept logit:
\begin{equation}
  e_{i,c}(p)=\bm{w}_c^{\top}\bm{F}_i(p)+b_c,
  ~
  z_{i,c}=\frac{1}{|\Omega_F|}\sum_{p\in\Omega_F}e_{i,c}(p).
  \label{eq:eacbm_concept_response}
\end{equation}
The resulting bottleneck \(\bm{z}_i=[z_{i,1},\cdots,z_{i,m}]\) contains one response per named concept.
Image-level concept targets supervise the pooled logits, while region annotations guide the same filters and spatial features before pooling.
Spatial supervision thus acts on the maps generating the classifier's concept inputs.

Average pooling also provides an explicit additive decomposition of each concept response over spatial locations.
Since \(\partial z_{i,c} / \partial \bm{F}_i(p) = \bm{w}_c/|\Omega_F|\), so Grad-CAM~\cite{selvaraju2017grad} for $z_{i,c}$ at \(\bm{F}_i\) reduces to the positive part of \(e_{i,c}-b_c\), up to a positive scale.
The full evidence map determines the concept logit, while Grad-CAM visualizes its positive, feature-dependent component.
This connection makes the spatial evidence inspectable along the prediction path; the objectives in Section~\ref{sec:eacbm:training_obj} further train its alignment with, and dependence on, the corresponding input regions.

\subsection{Concept-to-Label Prediction}
\label{sec:eacbm:prediction}

To expose how concepts contribute to classification, we map the bottleneck to class scores with a dense linear classifier (Figure~\ref{fig:eacbm_architecture}, right).
We standardize each response using its training-set mean $\mu_c$ and standard deviation $\sigma_c$, placing concepts on comparable scales.
These statistics stay fixed at inference:
\begin{equation}
  s_{i,y}=\sum_{c=1}^{m} a_{y,c}\,\tilde{z}_{i,c}+d_y,
  \quad
  \tilde{z}_{i,c}=\frac{z_{i,c}-\mu_c}{\sigma_c},
  \label{eq:eacbm_classifier}
\end{equation}
The class-specific weights $a_{y,c}$ and biases $d_y$ are learned from class labels.
Each score decomposes into signed concept contributions \(a_{y,c}\tilde{z}_{i,c}\) and a class bias $d_y$, while the pre-pooling maps retain the spatial evidence underlying each response.
At inference, the image follows the same path: spatial features, concept evidence maps, pooled and standardized responses, and class scores.

\subsection{Training Objectives}
\label{sec:eacbm:training_obj}

Concept and class supervision establish the semantic and predictive roles of the bottleneck; region supervision additionally guides the evidence supporting its responses.
We jointly train the mapping in Section~\ref{sec:eacbm:mapping} and the classifier in Section~\ref{sec:eacbm:prediction}, with region objectives only where annotations exist.

Each training image \(x_i\) has a class label \(y_i\), binary targets \(t_{i,c} \in \{0,1\}\) over \(m\) named concepts, and validity indicators \(v_{i,c} \in \{0,1\}\) marking observed targets. 
Its mask set \(\mathcal{M}_i\) contains binary region annotations for positive concepts whose regions are annotated and visible.
Masks may be available for only a subset \(\mathcal{S}\) of training images, with \(\mathcal{M}_i = \varnothing\) for \(i \notin \mathcal{S}\).
The three region objectives in Sections~\ref{sec:eacbm_match}--\ref{sec:eacbm_perturb} therefore operate on images with masks (Figure~\ref{fig:eacbm_architecture}, bottom), while Section~\ref{sec:eacbm_optimization} combines them with concept and class losses.
This formulation accommodates masks on any fraction of training images (Section~\ref{sec:expt:fewmasks}).

For each image with masks, we sample one annotated positive concept per iteration with probability proportional to the inverse square root of its training frequency, favoring rare concepts.
We resample its mask onto \(\Omega_F\) by area averaging.
Let \(\mathcal{P}_{\mathrm{reg}}\) denote the sampled image-concept pairs and \(\alpha_{i,c} \in [0,1]\) their resampled mask values.
We write \(\langle u\rangle_{\alpha}=\sum_p\alpha(p)u(p)/\sum_p\alpha(p)\) for a mask-weighted spatial mean.
The operator \(\mathbb{E}_{\mathrm{reg}}^{\omega}\) averages region-pair losses using positive concept-balancing weights \(\omega_c\), normalized to sum to one over the sampled pairs.

\subsubsection{Region-Filter Matching}
\label{sec:eacbm_match}

Image-level supervision alone can associate a concept with other parts that frequently co-occur with it.
Region-filter matching addresses this ambiguity by using the annotated region's features to identify the corresponding concept filter (Figure~\ref{fig:eacbm_architecture}, bottom left).

We pool these features into \(\bm{r}_{i,c}=\langle\bm{F}_i\rangle_{\alpha_{i,c}}\) and compare them with the same filters that generate the evidence maps in Section~\ref{sec:eacbm:mapping}.
With hats denoting \(\ell_2\) normalization and \(\mathcal{V}_{i,c}=\{d:v_{i,d} \!=\! 1\} \cup \{c\}\), we minimize a softmax loss over cosine similarities:
\begin{equation}
  \mathcal{L}_{\mathrm{match}}
  =
  -\mathbb{E}_{\mathrm{reg}}^{\omega}\!\Bigg[
    \log\frac{\exp(\widehat{\bm{r}}_{i,c}^{\top}\widehat{\bm{w}}_c/\tau_m)}{\sum_{d\in\mathcal{V}_{i,c}}\exp(\widehat{\bm{r}}_{i,c}^{\top}\widehat{\bm{w}}_d/\tau_m)}
  \Bigg].
  \label{eq:eacbm_match}
\end{equation}
Here,  \(\tau_m\) controls the softmax temperature.
The loss encourages the region feature to match its own concept filter over the other observed concepts, providing a direct association between region appearance and concept identity.

\subsubsection{Spatial Concept Grounding}
\label{sec:eacbm_grounding}

Matching a pooled region feature to its concept filter does not prevent that filter from activating elsewhere.
Spatial concept grounding supervises the spatial layout of evidence (Figure~\ref{fig:eacbm_architecture}, bottom middle).

We use the bias-free map \(\widetilde{e}_{i,c} = e_{i,c}-b_c\) from Eq.~\eqref{eq:eacbm_concept_response} to isolate feature-dependent evidence, encouraging positive responses inside the annotated region and suppressing them outside:
\begin{equation}
    \begin{aligned}
    \mathcal{L}_{\mathrm{grnd}} = 
    &\tfrac{1}{2} \mathbb{E}_{\mathrm{reg}}^{\omega}\!\bigl[\langle\operatorname{softplus}(-\widetilde{e}_{i,c}/\tau_g)\rangle_{\alpha_{i,c}} \\
    &+\langle\operatorname{softplus}(\widetilde{e}_{i,c}/\tau_g)\rangle_{1-\alpha_{i,c}}\bigr].
    \end{aligned}
    \label{eq:eacbm_grounding_loss}
\end{equation}
The separately normalized spatial means weight foreground and background equally regardless of region size, and \(\tau_g\) controls the evidence temperature.

\subsubsection{Matched Regional Perturbations}
\label{sec:eacbm_perturb}

Spatial alignment alone does not ensure dependence on the intended region, because features at a location may encode information from elsewhere in the image.
Moreover, perturbing any region can alter a response through general image disruption.
Matched regional perturbations therefore compare target-region perturbations with non-target controls in the same image, encouraging a larger response decrease for the target while limiting changes under controls (Figure~\ref{fig:eacbm_architecture}, bottom right).

Both target and control perturbations replace the selected region with a feathered mixture of blurred image content, column means, and noise (Appendix~\ref{sec:appendix:impl}).
Controls are other annotated parts of the same image that are disjoint from the target and closest to it in area.
We supplement them with shifted copies of the target mask that lie at least 95\% inside the object and are disjoint from the target and one another.
For target- and control-perturbed images \(x_i^{\mathrm{tar}}\) and \(x_i^{\mathrm{ctrl},j}\), we define the signed changes in the concept logit:
\begin{equation}
    \begin{aligned}
    \Delta^{\mathrm{tar}}_{i,c}
    &=z_c(x_i^{\mathrm{tar}})-z_c(x_i),\\
    \Delta^{\mathrm{ctrl},j}_{i,c}
    &=z_c(x_i^{\mathrm{ctrl},j})-z_c(x_i).
    \end{aligned}
\end{equation}
Negative changes indicate response decreases.
Let \(\mathcal{P}_{\mathrm{int}}\subseteq\mathcal{P}_{\mathrm{reg}}\) contain pairs with at least two non-overlapping same-image controls, and let \(\mathcal{T}_{\mathrm{int}}\) contain their pair-control triples \((i,c,j)\).
Using uniform averages over these sets, we minimize:
\begin{equation}
  \begin{split}
    \raisetag{18pt}
    \mathcal{L}_{\mathrm{pert}}
    &=\mathbb{E}_{\mathcal{T}_{\mathrm{int}}}\!\left[\rho_{\delta}(\Delta^{\mathrm{ctrl},j}-\Delta^{\mathrm{tar}})\right]\\[-2pt]
    &\quad+\tfrac12\mathbb{E}_{\mathcal{P}_{\mathrm{int}}}\![\rho_{\delta}(-\Delta^{\mathrm{tar}})]
    \!+\!\tfrac12\mathbb{E}_{\mathcal{T}_{\mathrm{int}}}\![\psi(\Delta^{\mathrm{ctrl},j})],
  \end{split}
  \label{eq:eacbm_intervention_loss}
\end{equation}
where \(\rho_{\delta}(u) \!=\! \operatorname{softplus}(\delta - u)\), \(\psi(t) \!=\! \operatorname{smoothL1}(t,0)\), and pair indices are omitted for readability.
The first term encourages a larger decrease under target perturbation than under control perturbation. The second explicitly suppresses the target-perturbed response relative to the original, while the third keeps control-perturbed responses close to the original.
Together, these terms favor selective response suppression rather than separation achieved merely by increasing control responses.

\subsubsection{Joint Training Objective}
\label{sec:eacbm_optimization}

The region objectives guide concept evidence, while concept and class losses maintain semantics and class discrimination.
For minibatch \(\mathcal{B}\), the concept loss is binary cross-entropy over observed targets:
\begin{equation}
  \mathcal{L}_{\mathrm{cnpt}}
  =
  \frac{\sum_{i\in\mathcal{B},c}v_{i,c}\, \ell_{\mathrm{BCE}}^{(c)}(z_{i,c},t_{i,c})}{\sum_{i\in\mathcal{B},c}v_{i,c}},
  \label{eq:eacbm_concept_loss}
\end{equation}
where \(\ell_{\mathrm{BCE}}^{(c)}\) uses the training-set positive-class weight for concept \(c\).
During joint training, the class loss \(\mathcal{L}_{\mathrm{cls}}\) is cross-entropy from a linear head on batch-normalized concept logits, as detailed in Appendix~\ref{sec:appendix:impl}.
We jointly optimize the image-to-concept mapping and classifier with:
\begin{equation}
    \begin{aligned}
    \mathcal{L}_{\mathrm{EA}}
    &=\mathcal{L}_{\mathrm{cnpt}}
    +\mathcal{L}_{\mathrm{cls}}
    +\lambda_{\mathrm{match}}\cdot\mathcal{L}_{\mathrm{match}}\\[-2pt]
    &\quad+\lambda_{\mathrm{grnd}}\cdot\mathcal{L}_{\mathrm{grnd}}
    +\lambda_{\mathrm{pert}}\cdot\mathcal{L}_{\mathrm{pert}},
    \end{aligned}
    \label{eq:eacbm_total_loss}
\end{equation}

We first train for four warm-up epochs without region objectives, then increase their weights linearly over four epochs.
To balance loss scales, each region objective's base weight is calibrated once—at its first step with a nonzero gradient—and fixed thereafter:
\begin{equation}
  \lambda_u^{(0)}
  =
  \operatorname{clip}_{[10^{-4},10^2]}\!\Big(\frac{r_u \cdot g_{\mathrm{cnpt}}}{\max(g_u,\varepsilon)}\Big),
  \label{eq:eacbm_loss_weight_init}
\end{equation}
where \(g_{\mathrm{cnpt}}\) and \(g_u\) are loss-gradient norms over shared trainable parameters.
We set \(r_{\mathrm{match}} \!=\! 0.5\), \(r_{\mathrm{grnd}} \!=\! 1.0\), \(r_{\mathrm{pert}} \!=\! 0.25\), \(\tau_m \!=\! 0.12\), \(\tau_g \!=\! 0.20\), \(\delta \!=\! 0.25\), and sample up to four controls per pair.

After joint training, we freeze the image-to-concept mapping and refit the classifier by minimizing class prediction loss with the fixed training-set standardization in Eq.~\eqref{eq:eacbm_classifier}.
Region objectives apply only during training; inference requires one encoder pass, a \(1\times1\) convolution, average pooling, standardization, and a linear classifier. 
Further training settings and costs appear in Appendix~\ref{sec:appendix:impl}.

\section{Experiments}
\label{sec:expt}

We evaluate EA-CBM through five questions:
\begin{itemize}
  \item \textbf{RQ1: Predictive Performance.} Does evidence alignment preserve classification accuracy and concept prediction (Sections~\ref{sec:expt:classification} and~\ref{sec:expt:concept})?
  
  \item \textbf{RQ2: Spatial Alignment.} How accurately does EA-CBM localize concept evidence, particularly for small regions (Section~\ref{sec:expt:localization})?
  
  \item \textbf{RQ3: Annotation Efficiency.} How much concept and region supervision is needed for effective evidence alignment (Section~\ref{sec:expt:fewmasks})?
  
  \item \textbf{RQ4: Component Contributions.} How do the region objectives and encoder adaptation affect prediction, localization, and responses to region removal (Section~\ref{sec:expt:albation})?
  
  \item \textbf{RQ5: Visual Evidence.} What supporting evidence do attribution maps reveal for concept responses (Section~\ref{sec:expt:case})?
\end{itemize}

\subsection{Experimental Setup}
\label{sec:expt:setup}

\paragraph{Datasets}
We evaluate the performance on eight datasets: PACO-LVIS~\cite{ramanathan2023paco}, Pascal-Part~\cite{chen2014detect}, MonuMAI~\cite{lamas2021monumai}, FunnyBirds~\cite{hesse2023funnybirds}, CUB~\cite{wah2011caltech}, and CelebA~\cite{liu2015deep, lee2020maskgan} provide concept regions as masks, boxes, or part locations;
Flower102~\cite{nilsback2008automated} and Food101~\cite{bossard2014food} lack region annotations; each image inherits the class-level PS-CBM concept list \cite{zhao2026partially}, shared across all concept-supervised methods, and EA-CBM uses only concept and class losses.
We use CelebA under its research license following CEM's task~\cite{zarlenga2022concept}.
Appendix~\ref{sec:appendix:data} details dataset construction and splits.

\begin{table*}[!t]
\setlength{\aboverulesep}{0pt}\setlength{\belowrulesep}{0pt}
\centering
\small
\setlength{\tabcolsep}{4pt}
\renewcommand{\arraystretch}{1.25}
\caption{\textbf{Classification, concept prediction, and localization with CLIP ViT-L/14 (\%).} Abbreviations: LP, linear probe; LF, LF-CBM; PS, PS-CBM; VLG, VLG-CBM; Seq., sequential CBM; Joint, Joint CBM; Min., Minimal-CBM; PCBM, PCBM-CAV; SL, SL-CBM; Proto, ProtoCBM. Best values are \textbf{bold}; second-best are \underline{underlined}.
Dashes (--) denote inapplicable entries: LP has no concepts; Flower102 and Food101 lack regions for CTS, and PGCM requires part regions; label-free CBMs use their own vocabularies on these two datasets, so C-mAP against the class-level concept lists is omitted.
Appendix~\ref{sec:appendix:full} provides macro-F1, AUROC, and standard deviations.}
\label{tab:multidataset}
\vspace{-0.5em}
\resizebox{1.0\textwidth}{!}{%
\begin{tabular}{llcccccccccccccc}
\toprule
\rowcolor{tbhead} &  & \multicolumn{1}{c}{\textbf{Ref.}} & \multicolumn{4}{c}{\textbf{Label-Free}} & \multicolumn{5}{c}{\textbf{Concept-Supervised}} & \multicolumn{3}{c}{\textbf{Spatial}} & \multicolumn{1}{c}{\textbf{Ours}} \\
\headrulefill \cmidrule(lr){3-3} \cmidrule(lr){4-7} \cmidrule(lr){8-12} \cmidrule(lr){13-15} \cmidrule(lr){16-16}
\rowcolor{tbhead} \multirow{-2}{*}{\textbf{Dataset}} & \multirow{-2}{*}{\textbf{Metric}} & \textbf{LP} & \textbf{LF} & \textbf{LM4CV} & \textbf{PS} & \textbf{VLG} & \textbf{Seq.} & \textbf{Joint} & \textbf{CEM} & \textbf{Min.} & \textbf{PCBM} & \textbf{SL} & \textbf{Proto} & \textbf{PGCM} & \textbf{EA-CBM} \\
\midrule
& ACC $\uparrow$ & 85.5 & 81.7 & 83.0 & 85.0 & 78.1 & 82.8 & 79.0 & 81.6 & 83.0 & 83.7 & 87.9 & \secondbest{88.0} & 77.2 & \textbf{88.4}  \\
& C-mAP $\uparrow$ & -- & 10.7 & 10.5 & 49.8 & 37.0 & 57.9 & 46.2 & \secondbest{86.8} & 54.3 & 54.2 & 54.8 & 49.3 & 75.2 & \textbf{88.4}  \\
\multirow{-3}{*}{\textbf{PACO-LVIS}} & CTS $\uparrow$ & -- & 55.5 & \secondbest{56.0} & 47.9 & 30.3 & 52.0 & 45.1 & 52.7 & 46.8 & 51.7 & 43.4 & 32.7 & 46.2 & \textbf{83.2}  \\
\midrule
\rowcolor{tbzebra} & ACC $\uparrow$ & 95.0 & 91.4 & 91.1 & 94.2 & 94.3 & 93.0 & 89.2 & 94.6 & 94.7 & 93.0 & \secondbest{95.3} & \textbf{96.5} & 87.6 & \textbf{96.5}  \\
\rowcolor{tbzebra} & C-mAP $\uparrow$ & -- & 41.5 & 38.1 & 72.9 & 81.0 & 84.0 & 72.3 & \secondbest{86.1} & 85.6 & 78.7 & 82.8 & 81.6 & 85.7 & \textbf{90.2}  \\
\rowcolor{tbzebra} \multirow{-3}{*}{\textbf{Pascal-Part}} & CTS $\uparrow$ & -- & 57.6 & 58.6 & 53.8 & 56.5 & 55.7 & \secondbest{59.5} & 56.2 & 59.3 & 51.7 & 42.9 & 36.0 & 50.9 & \textbf{89.7}  \\
\midrule
& ACC $\uparrow$ & 88.2 & 72.6 & 71.9 & 83.9 & 87.1 & 81.4 & 83.0 & 86.2 & 86.4 & 82.0 & 84.9 & \textbf{92.8} & 88.5 & \secondbest{92.1}  \\
& C-mAP $\uparrow$ & -- & 21.3 & 20.8 & 37.2 & 39.9 & 56.9 & 50.7 & 67.7 & 71.0 & 63.1 & 70.0 & \secondbest{79.3} & 78.9 & \textbf{80.8}  \\
\multirow{-3}{*}{\textbf{MonuMAI}} & CTS $\uparrow$ & -- & 63.8 & 51.3 & 47.6 & 44.7 & \secondbest{64.5} & 49.5 & 62.4 & 58.5 & 60.3 & 52.4 & 61.8 & 62.2 & \textbf{91.0}  \\
\midrule
\rowcolor{tbzebra} & ACC $\uparrow$ & 78.8 & 19.8 & 28.9 & 80.4 & 81.0 & 64.1 & 81.3 & 84.5 & 85.5 & 58.9 & 81.4 & \textbf{94.9} & 81.3 & \secondbest{94.7}  \\
\rowcolor{tbzebra} & C-mAP $\uparrow$ & -- & 22.9 & 22.4 & 82.0 & 89.0 & 66.6 & 73.1 & \secondbest{95.4} & 93.4 & 57.4 & 80.1 & 72.3 & 82.1 & \textbf{99.3}  \\
\rowcolor{tbzebra} \multirow{-3}{*}{\textbf{FunnyBirds}} & CTS $\uparrow$ & -- & 66.5 & 72.8 & 61.7 & 61.8 & \secondbest{88.2} & 69.7 & 77.5 & 75.4 & 75.6 & 47.7 & 55.6 & 60.4 & \textbf{99.4}  \\
\midrule
& ACC $\uparrow$ & 84.4 & 76.1 & 73.4 & 82.0 & 78.1 & 79.7 & 57.1 & 79.6 & 61.4 & 82.5 & 68.0 & \secondbest{85.6} & 66.0 & \textbf{86.0}  \\
& C-mAP $\uparrow$ & -- & 18.2 & 18.7 & 27.8 & 19.5 & 28.2 & \secondbest{31.6} & 30.2 & 28.8 & 23.1 & 26.5 & 25.9 & 27.1 & \textbf{32.1}  \\
\multirow{-3}{*}{\textbf{CUB}} & CTS $\uparrow$ & -- & 34.3 & 32.0 & 34.3 & 22.4 & \secondbest{56.2} & 53.1 & 47.2 & 36.0 & 36.5 & 24.3 & 27.6 & 51.7 & \textbf{98.4}  \\
\midrule
\rowcolor{tbzebra} & ACC $\uparrow$ & 39.1 & 33.6 & 31.4 & \textbf{42.4} & 35.3 & 35.5 & 39.1 & 38.5 & 41.5 & 34.0 & 39.8 & \secondbest{42.2} & 38.6 & 40.9  \\
\rowcolor{tbzebra} & C-mAP $\uparrow$ & -- & 40.0 & 36.5 & 45.9 & 27.4 & 69.9 & 69.1 & \secondbest{72.9} & 71.1 & 57.9 & 68.5 & 42.1 & 69.2 & \textbf{73.2}  \\
\rowcolor{tbzebra} \multirow{-3}{*}{\textbf{CelebA}} & CTS $\uparrow$ & -- & \secondbest{91.3} & 83.2 & 78.7 & 17.5 & 85.9 & 86.4 & 86.8 & 85.8 & 79.4 & 66.7 & 62.9 & 73.0 & \textbf{99.9}  \\
\midrule
& ACC $\uparrow$ & \secondbest{99.4} & 97.9 & 98.2 & \secondbest{99.4} & \secondbest{99.4} & 98.4 & 94.7 & 98.5 & 98.7 & 98.9 & 98.8 & \textbf{99.5} & -- & 99.3  \\
\multirow{-2}{*}{\textbf{Flower102}} & C-mAP $\uparrow$ & -- & -- & -- & -- & -- & 79.3 & 68.1 & \secondbest{99.4} & 76.2 & 98.0 & 94.6 & 55.7 & -- & \textbf{99.5}  \\
\midrule

\rowcolor{tbzebra} & ACC $\uparrow$ & \textbf{93.1} & 91.7 & 91.8 & \textbf{93.1} & 92.0 & 92.4 & \secondbest{93.0} & 91.2 & 91.5 & 92.2 & 90.4 & 92.7 & -- & 92.5  \\
\rowcolor{tbzebra} \multirow{-2}{*}{\textbf{Food101}} & C-mAP $\uparrow$ & -- & -- & -- & -- & -- & 89.6 & \textbf{96.1} & \secondbest{95.8} & 91.6 & 88.8 & 90.0 & 4.5 & -- & 95.6  \\
\bottomrule
\end{tabular}}
\end{table*}

\paragraph{Baselines}
All methods share the CLIP ViT-L/14 backbone~\cite{radford2021learning} and data splits; results with CLIP RN50 appear in Appendix~\ref{sec:appendix:rn50}.
We compare against a linear probe; label-free CBMs LF-CBM~\cite{oikarinen2023label}, LM4CV~\cite{yan2023learning}, PS-CBM~\cite{zhao2026partially}, and VLG-CBM~\cite{srivastava2024vlg}; concept-supervised models Sequential and Joint CBM~\cite{koh2020concept}, CEM~\cite{zarlenga2022concept}, Minimal-CBM~\cite{almudevar2026never}, and PCBM-CAV~\cite{yuksekgonul2023post}; and spatial CBMs SL-CBM~\cite{zhang2026slcbm}, ProtoCBM~\cite{huang2024concept}, and PGCM~\cite{colamonaco2026prototype}.
Each baseline retains its official concept construction and objectives, with adaptations documented in Appendix~\ref{sec:appendix:baselines}.

\paragraph{Metrics}
We report classification accuracy (ACC), concept mean average precision (C-mAP) over observed test labels, and the Concept Trustworthiness Score (CTS)~\cite{huang2024concept}, adapted to annotated regions.
For each positive test image-concept pair with a visible region, we compute Grad-CAM~\cite{selvaraju2017grad} of the concept response at the method-specific spatial layer (Appendix~\ref{sec:appendix:baselines}).
A hit requires a \(90\times90\) window centered on the attribution maximum to intersect the region.
Hits are averaged within and then across concepts, counting omitted concepts as misses.
Metrics are percentages (\%); differences are percentage points. 
Main comparisons average three runs, except the linear probe (one run).

\paragraph{Implementation Details}
Optimization settings, training costs, and protocols are in Appendix~\ref{sec:appendix:impl}.

\subsection{Classification}
\label{sec:expt:classification}

We first assess whether evidence alignment preserves the predictive performance of the concept bottleneck and linear classifier (Section~\ref{sec:eacbm:prediction}).
In Table~\ref{tab:multidataset}, EA-CBM achieves the highest accuracy (ACC) on PACO-LVIS and CUB, ties ProtoCBM on Pascal-Part, and remains within 0.2--1.5 points of the best method on the remaining datasets.
It outperforms the linear probe with the same CLIP backbone by 1.5--15.9 points across all six datasets with concept regions.
The largest gain is on FunnyBirds, whose classes combine the part variants named by the concepts.

\subsection{Concept Prediction}
\label{sec:expt:concept}

We evaluate pooled concept responses to test whether spatial supervision preserves concept semantics.
EA-CBM achieves the highest C-mAP on all six datasets with regions, exceeding the strongest baseline by 0.3--4.1 points (Table~\ref{tab:multidataset}).
It reaches 88.4 on PACO-LVIS versus CEM's 86.8, with the largest gains on Pascal-Part (4.1) and FunnyBirds (3.9), whose concepts describe co-occurring object parts.
On Flower102 and Food101, C-mAP measures agreement with class-level concept lists (approximately 1\% average concept prevalence), distinct from concept prediction with region supervision.
With only concept and class losses, EA-CBM leads on Flower102 (99.5) and reaches 95.6 on Food101, 0.5 points below Joint CBM.

\subsection{Concept Localization}
\label{sec:expt:localization}

We next evaluate alignment with annotated regions using CTS.
EA-CBM leads on all six datasets with regions, exceeding the strongest baseline by 8.6--42.2 points (Table~\ref{tab:multidataset}), with the largest gains on CUB, Pascal-Part, and PACO-LVIS.
These gains far exceed EA-CBM's maximum CTS standard deviation across runs of 0.8 points (Appendix~\ref{sec:appendix:full}).

Removing region objectives sharply reduces CTS without architectural changes, linking gains to region supervision (Section~\ref{sec:expt:albation}).
Other spatial models align less well: ProtoCBM and SL-CBM trail every concept-supervised baseline on 5 of 6 datasets; PGCM trails EA-CBM despite training its segmenter with region masks, supporting directly supervising evidence maps and pooled responses.

\begin{figure}[t]
\centering
\includegraphics[width=1.0\columnwidth]{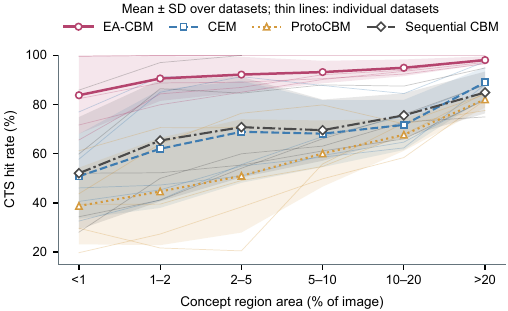}
\vspace{-1.5em}
\caption{\textbf{Localization by region size.}
CTS hit rates for positive test pairs grouped by concept-region area. EA-CBM is compared with baselines that most often lead in C-mAP (CEM), accuracy (ProtoCBM), and CTS (Sequential CBM).
Curves average PACO-LVIS, Pascal-Part, MonuMAI, FunnyBirds, and CelebA over nonempty bins; CUB is excluded because it provides part locations.
Bands show one standard deviation across datasets, and thin lines show individual datasets; seed-1 runs.}
\label{fig:area}
\vspace{-0.5em}
\end{figure}

Figure~\ref{fig:area} examines localization by region size, comparing EA-CBM with CEM, ProtoCBM, and Sequential CBM, the baselines most often leading in C-mAP, ACC, and CTS, respectively.
For regions below 1\% of image area, EA-CBM achieves an 83.8\% hit rate versus 38.8--52.1\% for these baselines.
Small regions tolerate less misplaced attribution, consistent with the largest overall gain on CUB, which annotates small parts including eyes, crowns, and beaks.

\subsection{Learning from Few Annotations}
\label{sec:expt:fewmasks}

\begin{figure}[t]
\centering
\includegraphics[width=1.0\columnwidth]{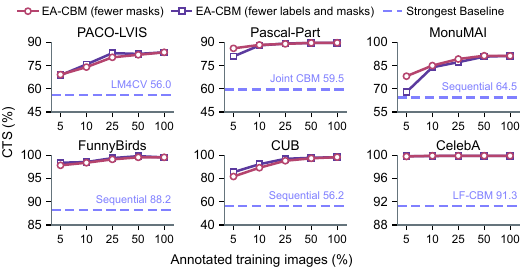}
\vspace{-1.5em}
\caption{\textbf{Localization with limited annotations.}
CTS (\%) on six datasets with masks alone (magenta) or concept labels and masks (purple) on 5--100\% of training images (100\%: full model).
Images without masks incur no region loss; those lacking concept labels receive only class supervision.
Dashed lines show the strongest baseline from Table~\ref{tab:multidataset}, with full concept supervision where applicable.}
\label{fig:maskfraction}
\vspace{-0.5em}
\end{figure}

EA-CBM supports partial region supervision by applying region losses only to images with masks (Section~\ref{sec:eacbm:training_obj}).
Figure~\ref{fig:maskfraction} compares annotation settings on nested random subsets of 5--50\% of training images.

We first restrict masks while retaining all concept and class labels.
Masks on just 5\% of images raise CTS by 28.9--64.6 over the model without region objectives and surpass the strongest baseline on every dataset (Table~\ref{tab:maskfraction}).
At 25\%, CTS is within 3.3 points of full region supervision.
Then, we restrict both concept labels and masks to the same subsets, leaving only class supervision elsewhere.
Validation annotations are restricted analogously; test annotations remain complete.
At 5\%, EA-CBM reaches CTS of 67.9--99.8 versus 13.4--62.4 for matched controls without region objectives (Table~\ref{tab:labelfraction}).
It also exceeds each dataset's strongest baseline under its standard supervision (Table~\ref{tab:multidataset}) by 3.4 (MonuMAI) to 29.3 (CUB).

In both settings, accuracy remains within 1.6 points of full supervision.
With fewer concept labels, C-mAP increases with the annotated fraction and consistently matches or exceeds corresponding controls on four datasets, by up to 4.1 points.
Region objectives thus improve localization even with concept labels and masks on few images.

\subsection{Ablation Study}
\label{sec:expt:albation}

\begin{table}[!t]
\setlength{\aboverulesep}{0pt}\setlength{\belowrulesep}{0pt}
\centering
\small
\setlength{\tabcolsep}{2.5pt}
\renewcommand{\arraystretch}{1.25}
\caption{\textbf{Objective ablation and controls} on six datasets with concept regions (\%, one run each). 
Leave-one-out columns omit one region objective; \(-\)All retains only concept and class losses; Frozen fixes the final encoder block. Best values in each row are bold, second-best underlined.}
\label{tab:ablation}
\vspace{-0.5em}
\resizebox{0.99\columnwidth}{!}{%
\begin{tabular}{llcccccc}
\toprule
\rowcolor{tbhead} \textbf{Dataset} & \textbf{Metric} & \textbf{Full} & \bm{$-\mathcal{L}_{\mathrm{match}}$} & \bm{$-\mathcal{L}_{\mathrm{grnd}}$} & \bm{$-\mathcal{L}_{\mathrm{pert}}$} & \bm{$-\mathrm{All}$} & \textbf{Frozen} \\
\midrule
& ACC $\uparrow$ & 88.3 & 88.4 & 88.4 & \secondbest{88.5} & \textbf{88.6} & 86.2 \\
& C-mAP $\uparrow$ & \secondbest{88.5} & 88.1 & \textbf{88.9} & 87.9 & 88.2 & 82.7 \\
\multirow{-3}{*}{\textbf{PACO-LVIS}} & CTS $\uparrow$ & \textbf{83.6} & 60.8 & 79.3 & \secondbest{82.9} & 39.9 & 81.3 \\
\midrule
\rowcolor{tbzebra} & ACC $\uparrow$ & 96.5 & \secondbest{96.6} & 96.5 & \secondbest{96.6} & \textbf{96.7} & 95.8 \\
\rowcolor{tbzebra} & C-mAP $\uparrow$ & \secondbest{90.1} & \textbf{90.2} & \textbf{90.2} & 89.6 & \secondbest{90.1} & 83.7 \\
\rowcolor{tbzebra} \multirow{-3}{*}{\textbf{Pascal-Part}} & CTS $\uparrow$ & \textbf{89.7} & 73.0 & 88.9 & \secondbest{89.4} & 35.6 & 87.9 \\
\midrule
& ACC $\uparrow$ & \textbf{93.5} & \secondbest{93.1} & \secondbest{93.1} & 91.9 & 91.9 & 87.2 \\
& C-mAP $\uparrow$ & \secondbest{81.5} & 80.5 & \textbf{82.7} & 77.7 & 79.0 & 60.0 \\
\multirow{-3}{*}{\textbf{MonuMAI}} & CTS $\uparrow$ & 91.3 & 81.9 & 90.4 & \textbf{92.4} & 49.2 & \secondbest{91.5} \\
\midrule
\rowcolor{tbzebra} & ACC $\uparrow$ & \secondbest{94.6} & \secondbest{94.6} & \textbf{95.2} & \secondbest{94.6} & \secondbest{94.6} & 81.8 \\
\rowcolor{tbzebra} & C-mAP $\uparrow$ & \textbf{99.4} & \textbf{99.4} & 99.0 & \secondbest{99.3} & 99.0 & 63.4 \\
\rowcolor{tbzebra} \multirow{-3}{*}{\textbf{FunnyBirds}} & CTS $\uparrow$ & \textbf{99.6} & 95.4 & \secondbest{98.9} & \textbf{99.6} & 33.2 & 95.7 \\
\midrule
& ACC $\uparrow$ & 85.9 & \textbf{86.3} & 85.7 & 86.1 & \secondbest{86.2} & 75.5 \\
& C-mAP $\uparrow$ & 32.0 & 31.2 & \textbf{33.0} & 32.1 & \secondbest{32.8} & 25.1 \\
\multirow{-3}{*}{\textbf{CUB}} & CTS $\uparrow$ & \textbf{98.4} & 91.1 & 97.1 & \secondbest{98.2} & 17.8 & 95.5 \\
\midrule

\rowcolor{tbzebra} & ACC $\uparrow$ & \secondbest{41.2} & 40.8 & 40.8 & 41.0 & 40.9 & \textbf{41.4} \\
\rowcolor{tbzebra} & C-mAP $\uparrow$ & 73.2 & 73.1 & \secondbest{73.3} & 73.2 & \textbf{73.4} & 64.8 \\
\rowcolor{tbzebra} \multirow{-3}{*}{\textbf{CelebA}} & CTS $\uparrow$ & \textbf{99.9} & 98.5 & \textbf{99.9} & \textbf{99.9} & 64.8 & \secondbest{99.8} \\
\bottomrule
\end{tabular}}
\end{table}

\begin{figure*}[t]
\centering
\includegraphics[width=1.0\textwidth]{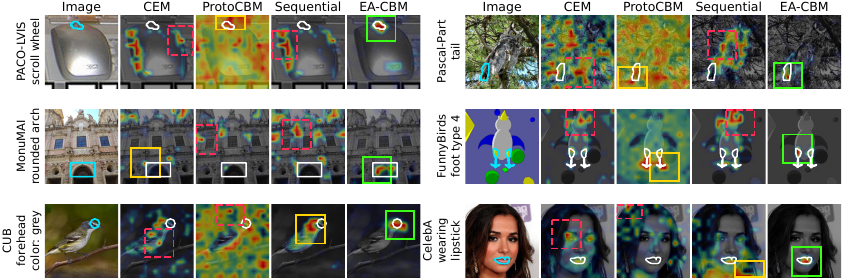}
\vspace{-1.75em}
\caption{\textbf{Concept attribution maps on test images}, one per dataset with concept regions, comparing EA-CBM with the baselines in Figure~\ref{fig:area}.
Boxes mark a \(90\times90\) CTS window centered on attribution peaks: green, peak inside the region; yellow, window intersects the region, but the peak lies outside; red, miss.
Contours mark annotated regions; circles mark CUB landmarks (green: peak within 14 pixels). Concepts are among the five most specific to each image's class; regions cover 1--15\% of the image away from its border.} 
\label{fig:case}
\vspace{-0.75em}
\end{figure*}

Table~\ref{tab:ablation} evaluates each region objective (Section~\ref{sec:eacbm:training_obj}) by removing it while retaining the architecture and remaining losses.
Additional controls remove all region objectives (\(-\text{All}\), retaining concept and class losses) or freeze the final encoder block.
Each setting keeps the training protocol and hyperparameters of the full model and uses one run.

\paragraph{Region Objectives and Localization}
Removing all region objectives reduces CTS by 35.1--80.6 points, with small changes in ACC and C-mAP, showing that concept prediction alone does not ensure spatial alignment.
Region-filter matching has the largest individual effect, reflecting its role in associating region features with the correct concept.
Spatial grounding contributes most on PACO-LVIS and CUB.
Each single-loss ablation substantially outperforms \(-\text{All}\) in CTS.

\paragraph{Adapting the Final Encoder Block}
Freezing the final encoder block while training the concept filters (Section~\ref{sec:eacbm:mapping}) keeps CTS within 3.9 points of the full model on every dataset.
Region supervision therefore remains effective for localization with frozen features, although freezing often reduces ACC and C-mAP (Table~\ref{tab:ablation}).

\paragraph{Matched Perturbations \& Concept Responses}
Matched perturbations target response dependence, which localization alone does not measure.
We evaluate concept-logit changes under a fixed test-time removal procedure distinct from the training perturbations (Appendix~\ref{sec:appendix:response}).
Mean target-region drops reach 0.20--3.26 test-set standard deviations with the full model, versus 0.10--0.84 without $\mathcal{L}_{\mathrm{pert}}$, while mean control-region drops remain at most 0.21 in both.
With this objective, target drops more often exceed mean control drops, supporting response selectivity (Section~\ref{sec:eacbm_perturb}).
Removing it also lowers ACC and C-mAP on MonuMAI, indicating benefits beyond localization.

\subsection{Case Study}
\label{sec:expt:case}

Figure~\ref{fig:case} illustrates the small-region localization gaps in Section~\ref{sec:expt:localization}.
Baseline attributions often peak on other parts or background: CEM and Sequential CBM highlight the head for the FunnyBirds foot, and the branch and body, respectively, for the Pascal-Part tail; for lipstick, CEM peaks on the cheek and ProtoCBM on the background.
For the PACO-LVIS scroll wheel, CEM and Sequential CBM peak on the sides of the mouse, and for the CUB forehead, CEM peaks below the head.
Nearby peaks can still count as CTS hits, as with CEM beside the rounded arch, Sequential CBM on the chin below the lips and beside the CUB forehead, and ProtoCBM near the scroll wheel, the Pascal-Part tail, and the FunnyBirds foot.
Every example includes such a yellow baseline box: the \(90\times90\) CTS window accepts nearby peaks, making CTS an upper bound on stricter peak-in-region hit rates.

EA-CBM meets the green criterion in all six examples, including the mouse's scroll wheel, with a 14-pixel landmark tolerance for CUB.
Its peaks also lie on the Pascal-Part tail, the rounded arch, the FunnyBirds feet, the CUB forehead, and the lips.
Its displayed attributions concentrate on target regions and remain weak elsewhere, whereas baseline maps often spread across the object and background.
This behavior is consistent with the complementary roles of region-filter matching and spatial concept grounding, which associate concepts with their regions and suppress off-region positive evidence.
These examples reflect its CTS of 83.2--99.9 across the six datasets with concept regions (Table~\ref{tab:multidataset}).

\section{Conclusions}
\label{sec:conclusions}

We introduced EA-CBM, an effective framework that aligns spatial evidence maps with annotated regions and pools them into the concept bottleneck used for classification.
Region-filter matching and spatial concept grounding localize evidence, while matched regional perturbations encourage larger response drops when the target region is removed than when a control region is removed.
Region annotations are required only during training.
Across six datasets with concept regions, EA-CBM achieves the highest CTS and C-mAP among evaluated methods, with classification accuracy comparable to concept-supervised CBMs.
With concept labels and masks on only 5\% of training images, it still surpasses every evaluated baseline in CTS.
These findings motivate joint training and evaluation of future CBMs for concept semantics, spatial alignment, and regional response selectivity.
Automatically generated region supervision could further reduce manual annotation.

{
  \small
  \bibliographystyle{ieeenat_fullname}
  \bibliography{main}
}
\setcounter{figure}{0}
\setcounter{table}{0}
\renewcommand{\thefigure}{A\arabic{figure}}
\renewcommand{\thetable}{A\arabic{table}}
\makeatletter\setlength{\@dblfptop}{0pt}\makeatother
\section*{Appendix}
\appendix

\section{Datasets}
\label{sec:appendix:data}

This section details the shared label and region formats, selection criteria, and construction of the eight datasets in Section~\ref{sec:expt:setup}.

\paragraph{Overview}
Table~\ref{tab:datasets} summarizes class and concept counts, splits, and region annotations.
All methods share images, class labels, and splits; test data are excluded from training and model selection.
Concept targets are positive, negative, or unobserved; EA-CBM's concept loss and test concept metrics use only observed targets.
Each concept region unites its annotated instances within an image, using part masks (PACO-LVIS, Pascal-Part, CelebA), expert boxes (MonuMAI), rendered part maps (FunnyBirds), or part locations (CUB).

Images and masks follow CLIP resizing and center cropping to \(224\times224\), with nearest-neighbor interpolation for masks; positives whose regions are cropped out become unobserved.
CTS uses all positive test image-concept pairs with visible regions, counted in the \emph{Pairs} column of Table~\ref{tab:datasets}.
For datasets with regions, we retain images, objects, classes, and concepts supporting region supervision and evaluation.
Selection uses only annotations, including object size, part counts, label frequency, and image provenance, with rules fixed before training and applied identically across methods.

\paragraph{PACO-LVIS \cite{ramanathan2023paco}}
PACO-LVIS contains 37 object classes and 133 generic-part concepts, using object-box crops with 10\% context.
Its federated labels are positive or negative for verified presence or absence, and unobserved otherwise.
To preserve parts after CLIP cropping, objects require at least three annotated parts, a box short side of at least 64 pixels, and minimum object and part mask areas of 400 and 25 pixels, respectively.
Parts require at least 50 positive objects in training and validation records and 10 in test; class thresholds are 100 and 20 objects, respectively.
Released training and validation records are pooled and resplit by source image; released test records are retained.

\paragraph{Pascal-Part \cite{chen2014detect}}
Pascal-Part contains 12 PASCAL VOC 2010~\cite{everingham2010pascal} object classes and 34 generic-part concepts, merging left, right, and numbered instances, \eg, both eyes form \emph{eye}.
Square crops have side length 1.2 times the object box's longer side, are centered on the box and shifted inside the image, and preserve the object through center cropping.
Eligible objects have at least three annotated parts and a box short side of at least 64 pixels.
Retained classes contain at least 100 eligible training objects; parts require annotations on at least 20 training objects.
Exhaustive part annotations make absent parts negatives.
VOC 2010 validation serves as the test set; validation holds out 10\% of training images, grouped by source image and stratified by class.

\paragraph{MonuMAI \cite{lamas2021monumai}}
MonuMAI labels building facades with four architectural styles and 15 architectural elements, each localized by the union of its expert bounding boxes.
Within each style, photos are grouped by capture time and file name into 700 groups, assigned intact (70/10/20) to training, validation, and test sets.

\begin{table}[t]
\setlength{\aboverulesep}{0pt}\setlength{\belowrulesep}{0pt}
\centering
\small
\setlength{\tabcolsep}{2pt}
\renewcommand{\arraystretch}{1.5}
\caption{\textbf{Statistics of the eight datasets.} 
Classes, concepts (concepts with region annotations in parentheses where not all), split sizes, positive test image-concept pairs with regions used for CTS, and the source of region annotations; Flower102 and Food101 have no regions, so they have no pairs (--).}
\label{tab:datasets}
\vspace{-0.5em}
\resizebox{0.99\columnwidth}{!}{%
\begin{tabular}{lrrrrrrr}
\toprule
\rowcolor{tbhead} \textbf{Dataset} & \textbf{\# Class} & \textbf{\# Concepts} & \textbf{Train} & \textbf{Val.} & \textbf{Test} & \textbf{Pairs} & \textbf{Regions} \\
\midrule

\textbf{PACO-LVIS} & 37 & 133 & 12,382 & 1,769 & 3,538 & 13,866 & Part Masks \\ 
\rowcolor{tbzebra} \textbf{Pascal-Part} & 12 & 34 & 5,206 & 526 & 5,894 & 45,010 & Part Masks \\ 
\textbf{MonuMAI} & 4 & 15 & 1,058 & 135 & 321 & 684 & Expert Boxes \\ 
\rowcolor{tbzebra} \textbf{FunnyBirds} & 50 & 26 & 45,000 & 5,000 & 500 & 2,411 & Rendered Parts \\ 
\textbf{CUB} & 200 & 312 (278) & 3,994 & 2,000 & 5,794 & 152,125 & Part Locations \\ 
\rowcolor{tbzebra} \textbf{CelebA} & 96 & 38 (35) & 23,748 & 2,929 & 2,772 & 22,744 & Face-Part Masks \\ 
\textbf{Flower102} & 102 & 1,068 (0) & 4,093 & 1,633 & 2,463 & -- & None \\ 
\rowcolor{tbzebra} \textbf{Food101} & 101 & 1,087 (0) & 50,500 & 20,200 & 30,300 & -- & None \\ 
\bottomrule
\end{tabular}}
\end{table}

\paragraph{FunnyBirds \cite{hesse2023funnybirds}}
FunnyBirds defines 50 synthetic bird classes through 26 part variants: four beak, three eye, four foot, nine tail, and six wing types.
Render parameters provide concept labels; rendered part maps provide regions with left and right parts merged.
The object region excludes background distractors.
Training renders with removed parts are retained, with removed variants labeled negative.
We hold out 10\% of training renders for validation and use the released 500-image test set.

\paragraph{CUB \cite{wah2011caltech}}
CUB contains 200 bird species and 312 per-image attributes.
Released training images are split into 3,994 training and 2,000 validation images; the 5,794 test images are retained.
Attribute groups map to their named parts, \eg, wing color to both wings and underparts color to the belly, breast, and throat.

Training regions are squares centered on part locations, with side lengths as fractions of the bird box's longer side: 0.12 for eyes; 0.18 for the beak, crown, forehead, nape, throat, and legs; 0.25 for the tail; 0.28 for the breast, belly, and back; and 0.32 for wings.
Test regions are annotated part locations, following ProtoCBM's CTS protocol.
Whole-bird attributes (size, shape, and primary color) remain in the concept loss without regions, leaving 278 concepts with regions (Table~\ref{tab:datasets}).

\paragraph{CelebA \cite{liu2015deep, lee2020maskgan}}
We follow CEM's task~\cite{zarlenga2022concept} using CelebAMask-HQ images and face-part masks with CelebA's training, validation, and test partitions.
Classes are binary patterns of eight attributes closest to 50\% positive: mouth slightly open, smiling, high cheekbones, heavy makeup, wearing lipstick, attractive, male, and arched eyebrows.
We retain the 96 patterns with at least 20 training images.
Excluding Attractive and Male from the concept set leaves 38 attributes.
Concept regions follow relevant face parts, \eg, lips for lipstick, glasses for eyeglasses, and hair for hair colors.
Whole-face attributes bald, blurry, and young remain in the concept loss without regions.

Both datasets are used under their non-commercial research licenses without redistribution; 
Figure~\ref{fig:case} illustrates attributions on one test image.
Released attributes reflect annotator judgments and dataset biases and serve only to benchmark concept localization.
We collect no new face data, recognize no identities, and release no face-attribute model for deployment.

\paragraph{Flower102~\cite{nilsback2008automated} and Food101~\cite{bossard2014food}}
We use PS-CBM's splits~\cite{zhao2026partially} and released class-level concept lists for Flower102 and Food101.
Each image inherits its class's list; the datasets contain 1,068 and 1,087 concepts, averaging 13.6 and 12.4 concepts per class, respectively.
Concept metrics therefore measure agreement with class-level lists.
Neither dataset provides regions, so EA-CBM uses only concept and class losses, and CTS is not computed.

\section{Implementation of EA-CBM}
\label{sec:appendix:impl}

This section details model training (Sections~\ref{sec:eacbm:mapping} and \ref{sec:eacbm:prediction}), regional perturbations (Section~\ref{sec:eacbm_perturb}), inference, and computational cost.

\paragraph{Image-to-Concept Mapping}
To reduce memory, the first 23 frozen CLIP ViT-L/14 blocks use half precision with float32 accumulation; the final block, concept filters, and losses use float32.
Concept filters operate on the \(16\times16\) grid of final-block patch tokens. 
Training images and their masks are flipped horizontally with probability 0.5.

We use AdamW with learning rates \(3\times10^{-3}\) for concept filters and \(2\times10^{-5}\) for the final block, weight decay \(10^{-4}\), batch size 128, cosine annealing, and at most 48 epochs.
The joint class loss has weight 1 and uses a linear head on batch-normalized concept logits, trained at the concept-filter learning rate.
Checkpoints minimize validation concept BCE, with higher C-mAP breaking ties; early stopping has a patience of 10 epochs.

\paragraph{Weights of the Region Objectives}
Region objectives start after four warm-up epochs, and their weights ramp linearly to the calibrated base values over four further epochs.
Each base weight in Eq.~\eqref{eq:eacbm_loss_weight_init} is calibrated when its objective first has a nonzero gradient and remains fixed thereafter.

\paragraph{Matched Perturbations}
Region-filter matching and spatial grounding reuse the current forward pass, whereas matched perturbations add one target and at most four control forward passes per eligible sampled image-concept pair.
We dilate each region mask by two pixels and feather it with a Gaussian of width 2. 
The fill combines a Gaussian-blurred image copy (\(\sigma=7\)) and its column means with weights 0.55 and 0.45, plus noise with standard deviation 0.08.

\paragraph{Concept-to-Label Classifier}
We fit the linear classifier on training-set standardized concept logits with unweighted cross-entropy, AdamW with learning rate \(3\times10^{-3}\), and batch size 256.
Five-fold cross-validation over the training and validation data selects weight decay from \(\{0,10^{-5},\dots,10^{-1}\}\) and training duration, up to 6,000 steps evaluated every 25 steps. 
This reduces reliance on small validation splits, such as MonuMAI's 135 images.
Classifier fitting leaves C-mAP and CTS unchanged because the concept model remains frozen.

\paragraph{Inference \& Cost}
Inference requires only the image: one encoder pass, \(1\times1\) concept filtering, average pooling, standardization, and linear classification.
A full run on one RTX 4090 takes 0.3 hours on MonuMAI and up to 4.6 hours on CelebA.
Main comparisons use three seeds, except the linear probe; the probe, ablation, limited-annotation, sensitivity, and region-removal studies use one.

\section{Baselines and Evaluation}
\label{sec:appendix:baselines}

This section describes the shared baseline protocol and method-specific CTS computation.

\paragraph{Baseline Protocol}
Each baseline retains its released concept construction, objectives, and concept-to-label predictor. 
All methods use the CLIP ViT-L/14 backbone and identical data splits.
ProtoCBM and PGCM are adapted from their original backbones; the linear probe fits a linear classifier to global CLIP image features and is run once.

\paragraph{CTS Evaluation}
For each concept response, Grad-CAM weights feature channels by their spatially averaged gradients, sums the weighted channels, applies ReLU, and bilinearly resizes the map to \(224\times224\).
A positive test pair with a visible region is a hit when the \(90\times90\) window (\(\pm45\) pixels) centered on the attribution maximum intersects that region; for CUB, it must contain the annotated part location.

We compute Grad-CAM at the final-block patch features for SL-CBM, ProtoCBM, and PGCM, and at the final-block input, the nearest spatial layer, for methods that read only the global image feature and for SL-CBM on PACO-LVIS.
Because \(z_c\) averages \(e_c\), EA-CBM's map is the positive part of its bias-free evidence map.
Hits are averaged within and then across concepts.
Omitted concepts receive constant scores for concept metrics and count as CTS misses.

\section{Further Localization Results}
\label{sec:appendix:localization}

We examine patterns hidden by aggregate CTS through region-size analysis and comparison of per-concept and per-pair averaging.

\paragraph{Region Size per Dataset}
Table~\ref{tab:area:perdataset} details Figure~\ref{fig:area} for the five datasets with region areas; CUB is excluded because its test annotations are part locations.
Pairs are binned at 1\%, 2\%, 5\%, 10\%, and 20\% of image area, and Figure~\ref{fig:area} averages each bin over datasets with nonempty entries; FunnyBirds has no regions above 5\%, leaving its last three bins empty.
EA-CBM exceeds every compared baseline in every nonempty bin.
Its lead over the strongest compared baseline is 2.5--11.9 points for regions above 20\% of image area and 13.7--39.9 points below 1\% (FunnyBirds and CelebA, respectively).
In the smallest bin, EA-CBM still achieves hit rates of 65.6--99.8\%.
These results show particularly large gains on small parts, consistent with the overall CTS advantage in Table~\ref{tab:multidataset}.

\begin{table}[!t]
\setlength{\aboverulesep}{0pt}\setlength{\belowrulesep}{0pt}
\centering
\small
\setlength{\tabcolsep}{2pt}
\renewcommand{\arraystretch}{1.4}
\caption{\textbf{Localization by region size on each dataset.} Share of CTS hits (\%) of EA-CBM and of the baselines in Figure~\ref{fig:area}, which lead most often in concept mAP (CEM), accuracy (ProtoCBM), and CTS (Sequential CBM), on the five datasets whose regions have an area; positive test pairs are grouped by the area of the concept region (\% of the image). Dashes (--) mark empty bins: FunnyBirds has no region larger than 5\% of the image. The largest value of each dataset and bin is bold; seed-1 runs.}
\label{tab:area:perdataset}
\vspace{-0.5em}
\resizebox{0.99\columnwidth}{!}{%
\begin{tabular}{llrrrrrr}
\toprule
\rowcolor{tbhead} \textbf{Dataset} & \textbf{Method} & \textbf{$<$1} & \textbf{1--2} & \textbf{2--5} & \textbf{5--10} & \textbf{10--20} & \textbf{$>$20} \\
\midrule
& Sequential CBM & 34.4 & 41.1 & 54.1 & 66.1 & 76.0 & 88.0 \\
& CEM & 32.6 & 41.3 & 55.6 & 67.0 & 75.3 & 88.1 \\
& ProtoCBM & 19.8 & 27.4 & 38.4 & 49.8 & 58.5 & 81.9 \\
\multirow{-4}{*}{\textbf{PACO-LVIS}} & \textbf{EA-CBM} & \textbf{71.7} & \textbf{80.0} & \textbf{85.0} & \textbf{90.2} & \textbf{93.5} & \textbf{97.2}  \\
\midrule
\rowcolor{tbzebra} & Sequential CBM & 52.1 & 52.3 & 55.4 & 61.2 & 66.0 & 81.9 \\
\rowcolor{tbzebra} & CEM & 46.1 & 47.3 & 51.0 & 59.3 & 64.8 & 82.4 \\
\rowcolor{tbzebra} & ProtoCBM & 39.4 & 39.5 & 49.1 & 54.9 & 67.8 & 84.7 \\
\rowcolor{tbzebra} \multirow{-4}{*}{\textbf{Pascal-Part}} & \textbf{EA-CBM} & \textbf{82.4} & \textbf{88.8} & \textbf{89.1} & \textbf{90.6} & \textbf{92.4} & \textbf{96.6}  \\
\midrule
& Sequential CBM & 28.1 & 50.0 & 60.0 & 63.1 & 72.9 & 75.0 \\
& CEM & 40.6 & 45.3 & 55.0 & 58.7 & 62.4 & 88.2 \\
& ProtoCBM & 43.8 & 64.1 & 76.5 & 79.9 & 72.2 & 77.6 \\
\multirow{-4}{*}{\textbf{MonuMAI}} & \textbf{EA-CBM} & \textbf{65.6} & \textbf{84.4} & \textbf{87.0} & \textbf{92.2} & \textbf{94.0} & \textbf{98.7}  \\
\midrule
\rowcolor{tbzebra} & Sequential CBM & 85.9 & 97.1 & \textbf{100.0} & -- & -- & -- \\
\rowcolor{tbzebra} & CEM & 77.0 & 91.3 & 91.8 & -- & -- & -- \\
\rowcolor{tbzebra} & ProtoCBM & 61.2 & 70.7 & 70.6 & -- & -- & -- \\
\rowcolor{tbzebra} \multirow{-4}{*}{\textbf{FunnyBirds}} & \textbf{EA-CBM} & \textbf{99.6} & \textbf{100.0} & \textbf{100.0} & -- & -- & --  \\
\midrule

& Sequential CBM & 59.9 & 86.5 & 84.7 & 88.0 & 87.5 & 94.9 \\
& CEM & 57.8 & 85.0 & 91.4 & 87.8 & 84.5 & 97.5 \\
& ProtoCBM & 29.7 & 21.7 & 20.5 & 55.9 & 72.5 & 84.2 \\
\multirow{-4}{*}{\textbf{CelebA}} & \textbf{EA-CBM} & \textbf{99.8} & \textbf{100.0} & \textbf{100.0} & \textbf{100.0} & \textbf{100.0} & \textbf{100.0}  \\
\bottomrule
\end{tabular}}
\end{table}

\paragraph{CTS over Test Pairs}
Per-concept average in Table~\ref{tab:multidataset} weighs rare and frequent concepts equally.
Table~\ref{tab:cts:micro} instead averages all positive test pairs, giving greater weight to frequent concepts.
EA-CBM again leads on every dataset by 10.3--41.5 points (CelebA and CUB), confirming its advantage under both averaging schemes; on PACO-LVIS, it reaches 86.6 versus 62.5 for LM4CV, the strongest baseline, a lead of 24.1 points.
Pair averaging raises 48 of the 72 baseline entries by up to 16.6 points (Joint CBM on PACO-LVIS) and lowers 21 by up to 6.5 (VLG-CBM on CelebA), whereas EA-CBM changes by at most 3.4 points on any dataset, so its localization holds for rare and frequent concepts alike.

\begin{table}[!t]
\setlength{\aboverulesep}{0pt}\setlength{\belowrulesep}{0pt}
\centering
\small
\setlength{\tabcolsep}{2pt}
\renewcommand{\arraystretch}{1.5}
\caption{\textbf{CTS averaged over test pairs} (\%) for the methods of Table~\ref{tab:multidataset} on the six datasets with concept regions. Each positive test image-concept pair counts once, so frequent concepts weigh more than in the per-concept average of Table~\ref{tab:multidataset}. Means over three runs; the largest value is bold and the second largest is underlined.}
\label{tab:cts:micro}
\vspace{-0.5em}
\resizebox{0.99\columnwidth}{!}{%
\begin{tabular}{lcccccc}
\toprule
\rowcolor{tbhead} \textbf{Method} & \textbf{PACO-LVIS} & \textbf{Pascal-Part} & \textbf{MonuMAI} & \textbf{FunnyBirds} & \textbf{CUB} & \textbf{CelebA} \\
\midrule
\textbf{LF-CBM} & 62.1 & 59.7 & \secondbest{69.0} & 66.0 & 31.4 & \secondbest{89.7} \\
\textbf{LM4CV} & \secondbest{62.5} & 57.6 & 64.5 & 71.1 & 33.0 & 83.4 \\
\textbf{PS-CBM} & 58.2 & 61.6 & 60.1 & 62.4 & 37.0 & 84.9 \\
\textbf{VLG-CBM} & 41.4 & 63.7 & 56.2 & 62.9 & 25.9 & 11.0 \\
\midrule
\rowcolor{tbzebra} \textbf{Sequential CBM} & 60.9 & 56.8 & 62.8 & \secondbest{87.7} & 55.8 & 85.5 \\
\rowcolor{tbzebra} \textbf{Joint CBM} & 61.7 & \secondbest{67.4} & 58.6 & 78.3 & 54.6 & 86.1 \\
\rowcolor{tbzebra} \textbf{CEM} & 60.8 & 56.2 & 63.1 & 78.7 & 47.8 & 86.5 \\
\rowcolor{tbzebra} \textbf{Minimal-CBM} & 58.5 & 64.1 & 60.9 & 75.6 & 39.7 & 85.7 \\
\rowcolor{tbzebra} \textbf{PCBM-CAV} & 59.5 & 53.6 & 60.3 & 75.7 & 36.5 & 77.7 \\
\midrule
\textbf{SL-CBM} & 51.5 & 40.0 & 50.2 & 48.5 & 24.7 & 64.0 \\
\textbf{ProtoCBM} & 46.6 & 40.9 & 68.4 & 59.0 & 35.9 & 57.2 \\
\textbf{PGCM} & 52.4 & 47.6 & 62.9 & 60.3 & \secondbest{56.8} & 69.3 \\
\midrule

\rowcolor{tbzebra} \textbf{EA-CBM} & \textbf{86.6} & \textbf{87.0} & \textbf{89.6} & \textbf{99.5} & \textbf{98.3} & \textbf{100.0}  \\
\bottomrule
\end{tabular}}
\end{table}

\section{Training with Limited Annotations}
\label{sec:appendix:partial}

This section details the annotation protocols and all three metrics for Section~\ref{sec:expt:fewmasks}.

\paragraph{Annotation Protocol}
We retain masks for nested random subsets of training images, sampled once with a fixed seed; images annotated at one fraction remain annotated at larger fractions.
Other images retain concept and class labels but contribute no region loss. 
The 0\% setting in Table~\ref{tab:maskfraction} is the \(-\)All control in Table~\ref{tab:ablation}.

\begin{table}[!t]
\setlength{\aboverulesep}{0pt}\setlength{\belowrulesep}{0pt}
\centering
\small
\setlength{\tabcolsep}{2pt}
\renewcommand{\arraystretch}{1.4}
\caption{\textbf{Training with region masks for a fraction of the training images} on the six datasets with concept regions. 0\% is the control without region objectives, and 100\% is the full model; images without masks get no region objective. ``Masked'' gives the number of training images with masks at 5\%. One run each. Higher is better; in each row, the largest value is bold, the second largest underlined.}
\label{tab:maskfraction}
\vspace{-0.5em}
\resizebox{0.99\columnwidth}{!}{%
\begin{tabular}{llccccccc}
\toprule
\rowcolor{tbhead} \textbf{Dataset} & \textbf{Metric} & \textbf{Masked} & \textbf{0\%} & \textbf{5\%} & \textbf{10\%} & \textbf{25\%} & \textbf{50\%} & \textbf{100\%} \\
\midrule
& ACC $\uparrow$ & 619 & \textbf{88.6} & 88.4 & 88.2 & 88.2 & \secondbest{88.5} & 88.3 \\
& C-mAP $\uparrow$ &  & 88.2 & 88.5 & \textbf{88.7} & \secondbest{88.6} & 88.5 & 88.5 \\
\multirow{-3}{*}{\textbf{PACO-LVIS}} & CTS $\uparrow$ &  & 39.9 & 69.2 & 74.0 & 80.3 & \secondbest{82.0} & \textbf{83.6} \\
\midrule
\rowcolor{tbzebra} & ACC $\uparrow$ & 260 & \textbf{96.7} & \secondbest{96.6} & 96.5 & 96.5 & 96.5 & 96.5 \\
\rowcolor{tbzebra} & C-mAP $\uparrow$ &  & \textbf{90.1} & 89.7 & 89.8 & \secondbest{90.0} & \textbf{90.1} & \textbf{90.1} \\
\rowcolor{tbzebra} \multirow{-3}{*}{\textbf{Pascal-Part}} & CTS $\uparrow$ &  & 35.6 & 86.1 & 88.3 & 89.0 & \secondbest{89.6} & \textbf{89.7} \\
\midrule
& ACC $\uparrow$ & 53 & 91.9 & 91.9 & 92.8 & 91.9 & \secondbest{93.1} & \textbf{93.5} \\
& C-mAP $\uparrow$ &  & 79.0 & 77.2 & 76.8 & 79.8 & \secondbest{80.0} & \textbf{81.5} \\
\multirow{-3}{*}{\textbf{MonuMAI}} & CTS $\uparrow$ &  & 49.2 & 78.1 & 85.0 & 89.2 & \textbf{91.4} & \secondbest{91.3} \\
\midrule
\rowcolor{tbzebra} & ACC $\uparrow$ & 2,250 & \secondbest{94.6} & \secondbest{94.6} & 94.2 & \secondbest{94.6} & \textbf{95.0} & \secondbest{94.6} \\
\rowcolor{tbzebra} & C-mAP $\uparrow$ &  & 99.0 & 99.2 & \secondbest{99.3} & \textbf{99.4} & \textbf{99.4} & \textbf{99.4} \\
\rowcolor{tbzebra} \multirow{-3}{*}{\textbf{FunnyBirds}} & CTS $\uparrow$ &  & 33.2 & 97.8 & 98.4 & \secondbest{99.1} & \textbf{99.6} & \textbf{99.6} \\
\midrule
& ACC $\uparrow$ & 200 & \secondbest{86.2} & \secondbest{86.2} & \textbf{86.4} & 86.1 & 85.8 & 85.9 \\
& C-mAP $\uparrow$ &  & \textbf{32.8} & 31.9 & 32.4 & \secondbest{32.5} & 32.1 & 32.0 \\
\multirow{-3}{*}{\textbf{CUB}} & CTS $\uparrow$ &  & 17.8 & 81.7 & 89.3 & 95.3 & \secondbest{97.4} & \textbf{98.4} \\
\midrule

\rowcolor{tbzebra} & ACC $\uparrow$ & 1,187 & 40.9 & 41.0 & 40.6 & \textbf{41.3} & 40.4 & \secondbest{41.2} \\
\rowcolor{tbzebra} & C-mAP $\uparrow$ &  & \textbf{73.4} & 73.2 & \secondbest{73.3} & \secondbest{73.3} & \secondbest{73.3} & 73.2 \\
\rowcolor{tbzebra} \multirow{-3}{*}{\textbf{CelebA}} & CTS $\uparrow$ &  & 64.8 & \secondbest{99.8} & \textbf{99.9} & \textbf{99.9} & \textbf{99.9} & \textbf{99.9} \\
\bottomrule
\end{tabular}}
\end{table}

\paragraph{Fewer Masks}
Table~\ref{tab:maskfraction} reports ACC, C-mAP, and CTS for Figure~\ref{fig:maskfraction}.
The largest gains occur when adding the first masks: increasing coverage from 0\% to 5\% raises CTS by 28.9--64.6 points, using only 53 masked images on MonuMAI and 200 on CUB.
With 5\% of the masks, EA-CBM exceeds each dataset's strongest baseline in Table~\ref{tab:multidataset} by 8.5--26.6 points (CelebA and Pascal-Part, respectively), reaching 83--100\% of full-model CTS.
At 25\%, CTS is within 3.3 points of full region supervision; from 50\% onward, it largely plateaus on Pascal-Part, MonuMAI, FunnyBirds, and CelebA.

Across fractions, ACC and C-mAP each vary by at most 0.9 points on five datasets.
On MonuMAI, the smallest dataset with 1,058 training images, their ranges are 1.6 and 4.7 points, respectively.

\paragraph{Fewer Concept Labels \& Masks}
Table~\ref{tab:labelfraction} retains concept labels and masks on the same nested 5--50\% subsets of training and validation images.
Training images outside these subsets contribute only the joint class loss; test annotations are fully retained.
The concept loss and its positive-class weights use only annotated training images, so \(-\)All receives the same concept supervision as EA-CBM.

\begin{table}[!t]
\setlength{\aboverulesep}{0pt}\setlength{\belowrulesep}{0pt}
\centering
\small
\setlength{\tabcolsep}{2pt}
\renewcommand{\arraystretch}{1.7}
\caption{\textbf{Training with concept labels and region masks for a fraction of the training images} on the six datasets with concept regions, for EA-CBM (EA) and the model without region objectives ($-$All). Both annotations are kept for the same images, and the other images keep only their class label; 100\% is the full model and $-$All is in Table~\ref{tab:ablation}. One run each. Higher is better; the larger value of each pair is bold.}
\label{tab:labelfraction}
\vspace{-0.5em}
\resizebox{0.99\columnwidth}{!}{%
\begin{tabular}{llcccccccccc}
\toprule
\rowcolor{tbhead} &  & \multicolumn{2}{c}{\textbf{5\%}} & \multicolumn{2}{c}{\textbf{10\%}} & \multicolumn{2}{c}{\textbf{25\%}} & \multicolumn{2}{c}{\textbf{50\%}} & \multicolumn{2}{c}{\textbf{100\%}} \\
\headrulefill \cmidrule(lr){3-4} \cmidrule(lr){5-6} \cmidrule(lr){7-8} \cmidrule(lr){9-10} \cmidrule(lr){11-12}
\rowcolor{tbhead} \multirow{-2}{*}{\textbf{Dataset}} & \multirow{-2}{*}{\textbf{Metric}} & \textbf{EA} & \textbf{$-$All} & \textbf{EA} & \textbf{$-$All} & \textbf{EA} & \textbf{$-$All} & \textbf{EA} & \textbf{$-$All} & \textbf{EA} & \textbf{$-$All} \\
\midrule

& ACC $\uparrow$ & \textbf{88.6} & 88.0 & 88.3 & \textbf{88.4} & 88.2 & \textbf{88.5} & 88.5 & 88.5 & 88.3 & \textbf{88.6} \\
& C-mAP $\uparrow$ & \textbf{78.8} & 76.1 & \textbf{84.4} & 82.6 & \textbf{87.0} & 86.1 & \textbf{88.0} & 87.7 & \textbf{88.5} & 88.2 \\
\multirow{-3}{*}{\textbf{PACO-LVIS}} & CTS $\uparrow$ & \textbf{68.8} & 39.9 & \textbf{75.8} & 40.0 & \textbf{83.0} & 40.3 & \textbf{82.5} & 40.5 & \textbf{83.6} & 39.9 \\
\midrule

\rowcolor{tbzebra} & ACC $\uparrow$ & \textbf{96.3} & 96.2 & \textbf{96.3} & 96.2 & \textbf{96.5} & 96.4 & 96.6 & 96.6 & 96.5 & \textbf{96.7} \\
\rowcolor{tbzebra} & C-mAP $\uparrow$ & \textbf{80.7} & 80.6 & \textbf{83.6} & 82.7 & \textbf{87.0} & 86.2 & \textbf{88.9} & 88.6 & 90.1 & 90.1 \\
\rowcolor{tbzebra} \multirow{-3}{*}{\textbf{Pascal-Part}} & CTS $\uparrow$ & \textbf{80.9} & 34.8 & \textbf{88.0} & 35.1 & \textbf{89.3} & 35.1 & \textbf{89.6} & 34.9 & \textbf{89.7} & 35.6 \\
\midrule

& ACC $\uparrow$ & \textbf{91.9} & 91.6 & \textbf{93.5} & 91.3 & \textbf{93.1} & 92.2 & \textbf{93.5} & 92.8 & \textbf{93.5} & 91.9 \\
& C-mAP $\uparrow$ & \textbf{47.2} & 44.3 & \textbf{57.1} & 53.0 & \textbf{68.3} & 64.9 & \textbf{77.4} & 75.0 & \textbf{81.5} & 79.0 \\
\multirow{-3}{*}{\textbf{MonuMAI}} & CTS $\uparrow$ & \textbf{67.9} & 48.2 & \textbf{83.9} & 53.8 & \textbf{87.2} & 49.4 & \textbf{90.8} & 51.5 & \textbf{91.3} & 49.2 \\
\midrule

\rowcolor{tbzebra} & ACC $\uparrow$ & 94.8 & \textbf{95.4} & 94.8 & \textbf{95.0} & 94.0 & \textbf{94.2} & \textbf{94.8} & 94.4 & 94.6 & 94.6 \\
\rowcolor{tbzebra} & C-mAP $\uparrow$ & 97.4 & 97.4 & 98.2 & 98.2 & \textbf{98.8} & 98.5 & \textbf{99.2} & 98.5 & \textbf{99.4} & 99.0 \\
\rowcolor{tbzebra} \multirow{-3}{*}{\textbf{FunnyBirds}} & CTS $\uparrow$ & \textbf{98.4} & 30.6 & \textbf{98.5} & 28.3 & \textbf{99.4} & 29.1 & \textbf{99.8} & 32.0 & \textbf{99.6} & 33.2 \\
\midrule

& ACC $\uparrow$ & 84.9 & \textbf{85.6} & \textbf{85.7} & 85.6 & 85.5 & \textbf{85.9} & 85.8 & \textbf{86.3} & 85.9 & \textbf{86.2} \\
& C-mAP $\uparrow$ & 23.0 & \textbf{23.1} & 25.2 & \textbf{25.3} & 29.1 & \textbf{29.2} & 30.5 & \textbf{31.2} & 32.0 & \textbf{32.8} \\
\multirow{-3}{*}{\textbf{CUB}} & CTS $\uparrow$ & \textbf{85.5} & 13.4 & \textbf{92.3} & 14.0 & \textbf{96.9} & 16.6 & \textbf{97.8} & 17.6 & \textbf{98.4} & 17.8 \\
\midrule

\rowcolor{tbzebra} & ACC $\uparrow$ & 41.9 & \textbf{42.0} & 40.0 & \textbf{42.2} & 40.8 & \textbf{41.3} & \textbf{41.3} & 40.1 & \textbf{41.2} & 40.9 \\
\rowcolor{tbzebra} & C-mAP $\uparrow$ & 63.4 & \textbf{63.5} & \textbf{67.4} & 66.3 & \textbf{70.8} & 69.1 & \textbf{72.3} & 71.8 & 73.2 & \textbf{73.4} \\
\rowcolor{tbzebra} \multirow{-3}{*}{\textbf{CelebA}} & CTS $\uparrow$ & \textbf{99.8} & 62.4 & \textbf{99.9} & 63.4 & \textbf{99.9} & 64.3 & \textbf{99.9} & 64.7 & \textbf{99.9} & 64.8 \\
\bottomrule
\end{tabular}}
\end{table}

With 5\% annotated images, e.g., 53 on MonuMAI and 200 on CUB, EA-CBM achieves CTS of 67.9--99.8, exceeding each dataset's strongest baseline under its standard supervision regime. 
From 10\% onward, CTS remains within 7.8 points of the full model.
By comparison, \(-\mathrm{All}\) stays within 13.4--64.8 across fractions, linking the localization gains to region supervision.
ACC remains within 1.6 points of full supervision, while C-mAP grows with the annotated fraction for both models.

\section{Sensitivity to Hyperparameters}
\label{sec:appendix:sensitivity}

We assess local sensitivity by independently halving and doubling the gradient ratios \(r_u\), temperatures \(\tau_m\) and \(\tau_g\), and margin \(\delta\) from Section~\ref{sec:eacbm_optimization}.

\paragraph{Evaluation Protocol}
We use one seed on MonuMAI, Pascal-Part, and CUB, the fastest datasets with regions, which cover expert boxes, part masks, and part locations.
The two runs of a hyperparameter thus differ by a factor of four, and all other settings, including the seed, splits, and classifier recipe, follow the full model.

\begin{table}[!t]
\setlength{\aboverulesep}{0pt}\setlength{\belowrulesep}{0pt}
\centering
\small
\setlength{\tabcolsep}{2.0pt}
\renewcommand{\arraystretch}{1.5}
\caption{\textbf{Sensitivity to the hyperparameters of the region objectives} on MonuMAI, Pascal-Part, and CUB (\%), which carry expert boxes, part masks, and part locations. Each row changes one hyperparameter of Section~\ref{sec:eacbm_optimization} and keeps the others at the default setting ($r_{\mathrm{match}}=0.5$, $r_{\mathrm{grnd}}=1.0$, $r_{\mathrm{pert}}=0.25$, $\tau_m=0.12$, $\tau_g=0.20$, $\delta=0.25$). One run each.}
\label{tab:sensitivity}
\vspace{-0.5em}
\resizebox{0.99\columnwidth}{!}{%
\begin{tabular}{lccccccccc}
\toprule
\rowcolor{tbhead} & \multicolumn{3}{c}{\textbf{MonuMAI}} & \multicolumn{3}{c}{\textbf{Pascal-Part}} & \multicolumn{3}{c}{\textbf{CUB}} \\
\headrulefill \cmidrule(lr){2-4} \cmidrule(lr){5-7} \cmidrule(lr){8-10}
\rowcolor{tbhead} \multirow{-2}{*}{\textbf{Setting}} & \textbf{ACC}$\uparrow$ & \textbf{C-mAP}$\uparrow$ & \textbf{CTS}$\uparrow$ & \textbf{ACC}$\uparrow$ & \textbf{C-mAP}$\uparrow$ & \textbf{CTS}$\uparrow$ & \textbf{ACC}$\uparrow$ & \textbf{C-mAP}$\uparrow$ & \textbf{CTS}$\uparrow$ \\
\midrule
\textbf{Default} & 93.5 & 81.5 & 91.3 & 96.5 & 90.1 & 89.7 & 85.9 & 32.0 & 98.4 \\
\rowcolor{tbzebra} \textbf{$\bm{r_{\mathrm{match}}}$=0.25} & 93.1 & 81.4 & 90.7 & 96.5 & 90.3 & 89.2 & 86.4 & 31.7 & 97.8 \\
\rowcolor{tbzebra} \textbf{$\bm{r_{\mathrm{match}}}$=1.0} & 92.8 & 81.7 & 92.3 & 96.6 & 89.8 & 90.1 & 85.7 & 32.1 & 98.6 \\
\textbf{$\bm{r_{\mathrm{grnd}}}$=0.5} & 93.5 & 82.2 & 91.1 & 96.5 & 90.2 & 89.6 & 85.8 & 32.4 & 98.2 \\
\textbf{$\bm{r_{\mathrm{grnd}}}$=2.0} & 93.5 & 80.3 & 91.7 & 96.5 & 89.9 & 89.6 & 86.0 & 31.4 & 98.5 \\
\rowcolor{tbzebra} \textbf{$\bm{r_{\mathrm{pert}}}$=0.125} & 92.2 & 80.0 & 91.9 & 96.6 & 90.0 & 89.7 & 86.1 & 31.9 & 98.3 \\
\rowcolor{tbzebra} \textbf{$\bm{r_{\mathrm{pert}}}$=0.5} & 93.5 & 82.2 & 91.1 & 96.5 & 90.2 & 89.9 & 86.0 & 32.0 & 98.4 \\
\textbf{$\bm{\tau_m}$=0.06} & 93.1 & 81.3 & 90.7 & 96.6 & 90.2 & 89.4 & 86.1 & 31.7 & 98.2 \\
\textbf{$\bm{\tau_m}$=0.24} & 93.1 & 81.5 & 91.4 & 96.5 & 89.8 & 89.1 & 85.5 & 31.9 & 98.5 \\
\rowcolor{tbzebra} \textbf{$\bm{\tau_g}$=0.10} & 93.5 & 81.5 & 91.3 & 96.5 & 90.1 & 89.7 & 86.0 & 32.0 & 98.4 \\
\rowcolor{tbzebra} \textbf{$\bm{\tau_g}$=0.40} & 93.5 & 81.5 & 91.6 & 96.5 & 90.1 & 89.7 & 85.9 & 32.0 & 98.4 \\
\textbf{$\bm{\delta}$=0.125} & 93.5 & 81.5 & 91.3 & 96.5 & 90.2 & 89.7 & 86.0 & 32.0 & 98.4 \\
\textbf{$\bm{\delta}$=0.5} & 93.5 & 81.5 & 91.3 & 96.5 & 90.1 & 89.7 & 85.9 & 32.0 & 98.4 \\
\bottomrule
\end{tabular}}
\end{table}

\paragraph{Sensitivity Results}
Across the twelve changes in Table~\ref{tab:sensitivity}, CTS remains within 1.0 point of the default setting, and ACC and C-mAP within 1.3 and 1.5 points, respectively.
ACC varies most on MonuMAI (92.2--93.5), where 1.3 points correspond to approximately four of 321 test images.
Region-filter matching has a consistent effect: doubling \(r_{\mathrm{match}}\) raises CTS on all three datasets, while halving it lowers CTS by 0.5--0.6 points.
Changing \(\tau_g\) and \(\delta\) affects each metric by at most 0.3 points, consistent with gradient-norm calibration limiting sensitivity to loss scale (Eq.~\eqref{eq:eacbm_loss_weight_init}).

\section{Region Objectives and Region Removal}
\label{sec:appendix:response}

We extend Table~\ref{tab:ablation} to all eight combinations of region-filter matching, spatial grounding, and matched perturbations, examining their effects on localization and regional response selectivity.
Region-removal evaluation also includes CEM and ProtoCBM, which most often lead the baselines in C-mAP and ACC, respectively, and PGCM, whose segmenter uses region masks.

\paragraph{Combination Protocol}
All settings retain the full training protocol and the concept and class losses, disabling omitted region objectives and using default gradient ratios \(r_u\) for active ones (Section~\ref{sec:eacbm_optimization}).
We reuse the zero-, two-, and three-objective runs from Table~\ref{tab:ablation} and add one run for each single-objective setting.
Thus, each setting uses one run, with the same Grad-CAM protocol as Table~\ref{tab:multidataset} for CTS.

\begin{table*}[!t]
\setlength{\aboverulesep}{0pt}\setlength{\belowrulesep}{0pt}
\centering
\small
\setlength{\tabcolsep}{1.5pt}
\renewcommand{\arraystretch}{1.6}
\caption{\textbf{Classification, concept prediction, and localization of EA-CBM trained with each combination of the region objectives} on the six datasets with concept regions (\%). Each row lists the region objectives used in training, and Full uses all three; all settings keep concept and class losses. One run each. Higher is better; in each column, the largest value is bold, the second largest underlined.}
\label{tab:combinations}
\vspace{-0.5em}
\resizebox{\textwidth}{!}{%
\begin{tabular}{lcccccccccccccccccc}
\toprule
\rowcolor{tbhead} & \multicolumn{3}{c}{\textbf{PACO-LVIS}} & \multicolumn{3}{c}{\textbf{Pascal-Part}} & \multicolumn{3}{c}{\textbf{MonuMAI}} & \multicolumn{3}{c}{\textbf{FunnyBirds}} & \multicolumn{3}{c}{\textbf{CUB}} & \multicolumn{3}{c}{\textbf{CelebA}} \\
\headrulefill \cmidrule(lr){2-4} \cmidrule(lr){5-7} \cmidrule(lr){8-10} \cmidrule(lr){11-13} \cmidrule(lr){14-16} \cmidrule(lr){17-19}
\rowcolor{tbhead} \multirow{-2}{*}{\textbf{Region Obj.}} & \textbf{ACC}$\uparrow$ & \textbf{C-mAP}$\uparrow$ & \textbf{CTS}$\uparrow$ & \textbf{ACC}$\uparrow$ & \textbf{C-mAP}$\uparrow$ & \textbf{CTS}$\uparrow$ & \textbf{ACC}$\uparrow$ & \textbf{C-mAP}$\uparrow$ & \textbf{CTS}$\uparrow$ & \textbf{ACC}$\uparrow$ & \textbf{C-mAP}$\uparrow$ & \textbf{CTS}$\uparrow$ & \textbf{ACC}$\uparrow$ & \textbf{C-mAP}$\uparrow$ & \textbf{CTS}$\uparrow$ & \textbf{ACC}$\uparrow$ & \textbf{C-mAP}$\uparrow$ & \textbf{CTS}$\uparrow$ \\
\midrule
\textbf{None} & \secondbest{88.6} & 88.2 & 39.9 & \textbf{96.7} & 90.1 & 35.6 & 91.9 & 79.0 & 49.2 & 94.6 & 99.0 & 33.2 & 86.2 & 32.8 & 17.8 & 40.9 & \secondbest{73.4} & 64.8 \\
\rowcolor{tbzebra} 
$\bm{\mathcal{L}_{\mathrm{match}}}$ & 88.5 & 87.9 & 77.4 & 96.5 & 89.5 & 88.4 & 91.9 & 79.6 & 90.6 & \textbf{95.6} & 99.2 & 98.8 & 86.0 & \textbf{33.0} & 96.5 & 40.4 & 73.3 & \textbf{99.9} \\
$\bm{\mathcal{L}_{\mathrm{grnd}}}$ & \textbf{88.8} & 87.7 & 60.6 & \secondbest{96.6} & 89.4 & 67.9 & 91.6 & 76.2 & 81.5 & 94.8 & \secondbest{99.3} & 94.8 & \textbf{86.5} & 31.4 & 86.5 & \secondbest{41.1} & 73.1 & \secondbest{98.5} \\
\rowcolor{tbzebra} 
$\bm{\mathcal{L}_{\mathrm{pert}}}$ & \secondbest{88.6} & \textbf{89.0} & 44.4 & \textbf{96.7} & \textbf{90.9} & 38.5 & \secondbest{93.1} & \secondbest{82.2} & 52.1 & 94.4 & \secondbest{99.3} & 34.9 & \textbf{86.5} & \secondbest{32.9} & 23.5 & 40.9 & \textbf{73.6} & 66.2 \\
$\bm{\mathcal{L}_{\mathrm{match}} \!+\! \mathcal{L}_{\mathrm{grnd}}}$ & 88.5 & 87.9 & \secondbest{82.9} & \secondbest{96.6} & 89.6 & \secondbest{89.4} & 91.9 & 77.7 & \textbf{92.4} & 94.6 & \secondbest{99.3} & \textbf{99.6} & 86.1 & 32.1 & \secondbest{98.2} & 41.0 & 73.2 & \textbf{99.9} \\
\rowcolor{tbzebra} 
$\bm{\mathcal{L}_{\mathrm{match}} \!+\! \mathcal{L}_{\mathrm{pert}}}$ & 88.4 & \secondbest{88.9} & 79.3 & 96.5 & \secondbest{90.2} & 88.9 & \secondbest{93.1} & \textbf{82.7} & 90.4 & \secondbest{95.2} & 99.0 & \secondbest{98.9} & 85.7 & \textbf{33.0} & 97.1 & 40.8 & 73.3 & \textbf{99.9} \\
$\bm{\mathcal{L}_{\mathrm{grnd}} \!+\! \mathcal{L}_{\mathrm{pert}}}$ & 88.4 & 88.1 & 60.8 & \secondbest{96.6} & \secondbest{90.2} & 73.0 & \secondbest{93.1} & 80.5 & 81.9 & 94.6 & \textbf{99.4} & 95.4 & \secondbest{86.3} & 31.2 & 91.1 & 40.8 & 73.1 & \secondbest{98.5} \\
\rowcolor{tbzebra} 
\textbf{Full} & 88.3 & 88.5 & \textbf{83.6} & 96.5 & 90.1 & \textbf{89.7} & \textbf{93.5} & 81.5 & \secondbest{91.3} & 94.6 & \textbf{99.4} & \textbf{99.6} & 85.9 & 32.0 & \textbf{98.4} & \textbf{41.2} & 73.2 & \textbf{99.9} \\
\bottomrule
\end{tabular}}
\end{table*}

\paragraph{Combination Results}
Table~\ref{tab:combinations} shows that region-filter matching accounts for most of the localization gain.
Matching alone achieves CTS of 77.4--99.9, exceeding \(-\mathrm{All}\) by 35.1--78.7 points and remaining within 6.2 points of the full model.
Spatial grounding alone improves CTS over \(-\mathrm{All}\) by 20.7--68.7 points; adding it to matching further improves localization on every dataset except CelebA.
Matched perturbations alone yield smaller CTS gains of 1.4--5.7 points.
The full model achieves the highest CTS on five of six datasets, including ties on FunnyBirds and CelebA.

Across all combinations, ACC varies by at most 1.9 points per dataset.
On MonuMAI, every variant with matched perturbations exceeds every variant without them in ACC (93.1--93.5 vs.\ 91.6--91.9) and C-mAP (80.5--82.7 vs.\ 76.2--79.6).
While matching and grounding primarily improve localization, matched perturbations are intended to affect response selectivity, which we evaluate next.

\paragraph{Region Removal Protocol}
To assess selectivity beyond general image disruption, we compare target-region removal with approximately area-matched removals elsewhere on the same object.
We sample up to 3,000 positive test image-concept pairs per dataset, dilate each target mask by 3 pixels, feather it with a width-3 Gaussian, and fill it with the mean color outside the mask.
This removal differs from training perturbations; CUB uses squares around part locations as during training.

Each target has up to two shifted-mask controls, each disjoint from the target, retaining at least 95\% of its area, and lying at least 95\% inside the object.
Evaluation uses pairs with valid controls, yielding 469--2,788 pairs per dataset.
All methods share the pairs, removals, and controls.
We evaluate concept logits throughout, converting CEM and PGCM probabilities to logits.

\begin{table*}[!t]
\setlength{\aboverulesep}{0pt}\setlength{\belowrulesep}{0pt}
\centering
\small
\setlength{\tabcolsep}{2pt}
\renewcommand{\arraystretch}{1.3}
\caption{\textbf{Concept response to region removal for CEM, ProtoCBM, PGCM, and EA-CBM trained with each combination of the region objectives} on the six datasets with concept regions. EA-CBM rows list the region objectives used in training, and all methods share the test pairs, removals, and controls. Drop Ratio (D-Ratio): mean target drop minus mean control drop, divided by the mean absolute control drop; Drop Rate (D-Rate): percentage of test pairs whose target drop exceeds their control drop. One run each. Higher is better; in each column, the largest value is bold, the second largest underlined.}
\label{tab:response}
\vspace{-0.5em}
\resizebox{\textwidth}{!}{%
\begin{tabular}{llllcccccccccccc}
\toprule
\rowcolor{tbhead} \multicolumn{2}{l}{} & \multicolumn{2}{c}{\textbf{PACO-LVIS}} & \multicolumn{2}{c}{\textbf{Pascal-Part}} & \multicolumn{2}{c}{\textbf{MonuMAI}} & \multicolumn{2}{c}{\textbf{FunnyBirds}} & \multicolumn{2}{c}{\textbf{CUB}} & \multicolumn{2}{c}{\textbf{CelebA}} \\
\headrulefill \cmidrule(lr){3-4} \cmidrule(lr){5-6} \cmidrule(lr){7-8} \cmidrule(lr){9-10} \cmidrule(lr){11-12} \cmidrule(lr){13-14}
\rowcolor{tbhead} \multicolumn{2}{l}{\multirow{-2}{*}{\textbf{Method}}} & \textbf{D-Ratio}$\uparrow$ & \textbf{D-Rate}$\uparrow$ & \textbf{D-Ratio}$\uparrow$ & \textbf{D-Rate}$\uparrow$ & \textbf{D-Ratio}$\uparrow$ & \textbf{D-Rate}$\uparrow$ & \textbf{D-Ratio}$\uparrow$ & \textbf{D-Rate}$\uparrow$ & \textbf{D-Ratio}$\uparrow$ & \textbf{D-Rate}$\uparrow$ & \textbf{D-Ratio}$\uparrow$ & \textbf{D-Rate}$\uparrow$ \\
\midrule
\multicolumn{2}{l}{\textbf{CEM}} & 0.24 & 51.3 & 0.40 & 62.0 & 0.68 & 69.3 & 3.08 & 82.8 & 0.64 & 67.7 & 0.20 & 53.8 \\
\rowcolor{tbzebra} 
\multicolumn{2}{l}{\textbf{ProtoCBM}} & 0.28 & 52.8 & 1.44 & 69.4 & \secondbest{7.94} & 82.1 & 7.19 & 92.6 & 0.99 & 60.2 & 2.21 & 70.5 \\
\multicolumn{2}{l}{\textbf{PGCM}} & \textbf{2.19} & 64.6 & \textbf{5.60} & 72.8 & \textbf{15.24} & 85.1 & 6.96 & 88.2 & \secondbest{4.76} & 83.7 & 13.48 & 92.1 \\
\midrule
 \rowcolor{tbzebra} & \textbf{None} & 0.47 & 57.1 & 1.97 & 78.2 & 2.37 & 82.3 & 4.74 & 92.9 & 1.43 & 66.3 & 2.38 & 76.1 \\
 \rowcolor{tbzebra} & $\bm{\mathcal{L}_{\mathrm{match}}}$ & 0.53 & 59.2 & 1.89 & 78.9 & 2.70 & 87.0 & 4.92 & 92.9 & 1.37 & 67.4 & 4.29 & 91.2 \\
 \rowcolor{tbzebra} & $\bm{\mathcal{L}_{\mathrm{grnd}}}$ & 0.52 & 58.5 & 1.83 & 77.8 & 1.73 & 77.6 & 5.00 & 93.5 & 1.20 & 65.6 & 2.84 & 80.2 \\
 \rowcolor{tbzebra} & $\bm{\mathcal{L}_{\mathrm{pert}}}$ & \secondbest{1.53} & 77.2 & \secondbest{4.36} & 93.8 & 7.62 & 94.9 & 7.27 & \textbf{96.8} & 4.65 & 95.2 & 14.17 & 99.5 \\
 \rowcolor{tbzebra} & $\bm{\mathcal{L}_{\mathrm{match}} \!+\! \mathcal{L}_{\mathrm{grnd}}}$ & 0.59 & 61.3 & 1.88 & 79.7 & 2.12 & 83.6 & 5.09 & 93.9 & 1.27 & 66.6 & 4.33 & 90.8 \\
 \rowcolor{tbzebra} & $\bm{\mathcal{L}_{\mathrm{match}} \!+\! \mathcal{L}_{\mathrm{pert}}}$ & 1.41 & 76.9 & \secondbest{4.36} & \textbf{94.4} & 7.40 & \secondbest{95.5} & 6.61 & 96.2 & 4.74 & \textbf{96.2} & 14.98 & \secondbest{99.9} \\
 \rowcolor{tbzebra} & $\bm{\mathcal{L}_{\mathrm{grnd}} \!+\! \mathcal{L}_{\mathrm{pert}}}$ & 1.41 & \textbf{78.7} & 3.97 & 93.5 & 7.39 & \textbf{95.7} & \secondbest{7.62} & \secondbest{96.5} & 4.54 & 95.4 & \secondbest{15.54} & 99.8 \\
\rowcolor{tbzebra} \multirow{-8}{*}{\rotatebox{90}{\textbf{EA-CBM}}} & \textbf{Full} & 1.39 & \secondbest{78.2} & 4.26 & \secondbest{94.2} & 7.66 & \secondbest{95.5} & \textbf{7.67} & 96.4 & \textbf{4.87} & \secondbest{96.1} & \textbf{17.38} & \textbf{100.0} \\
\bottomrule
\end{tabular}}
\end{table*}

\paragraph{Region Removal Metrics}
For each pair, \(d^{\mathrm{tar}}_{i,c}\) denotes the signed target-removal logit drop (original minus perturbed), and \(d^{\mathrm{ctrl}}_{i,c}\) the mean drop over its controls.
Both are normalized by the concept's test-set logit standard deviation.
With \(\mathbb{E}_{i,c}\) averaging evaluated pairs within each dataset, Table~\ref{tab:response} reports:
\begin{itemize}
  \item \textbf{Drop Ratio (D-Ratio)}: the mean target-control difference divided by the mean absolute control drop:
  \begin{equation}
    \text{D-Ratio}
    = \frac{\mathbb{E}_{i,c}[d^{\mathrm{tar}}_{i,c}-d^{\mathrm{ctrl}}_{i,c}]}
           {\mathbb{E}_{i,c}[\lvert d^{\mathrm{ctrl}}_{i,c}\rvert]}.
  \end{equation}
  \item \textbf{Drop Rate (D-Rate)}: the percentage of pairs whose target drop exceeds their mean control drop:
  \begin{equation}
    \text{D-Rate}
    = 100\% \times \mathbb{E}_{i,c}
      \!\left[\mathbf{1}\{d^{\mathrm{tar}}_{i,c}>d^{\mathrm{ctrl}}_{i,c}\}\right].
  \end{equation}
\end{itemize}

\paragraph{Region Removal Results}
Matched perturbations consistently improve regional response selectivity.
On each dataset, every combination containing this objective exceeds every combination without it in both metrics.
Across datasets and combinations, D-Ratio and D-Rate reach 1.39--17.38 and 76.9--100.0\% with matched perturbations, versus 0.47--5.09 and 57.1--93.9\% without them.
Adding matched perturbations to \(\mathcal{L}_{\mathrm{match}}+\mathcal{L}_{\mathrm{grnd}}\) increases D-Rate by 2.5--29.5 points and the mean target drop by factors ranging from 1.3 on FunnyBirds to 9.5 on CUB.
Mean control drops remain at most 0.22 test-set standard deviations across all eight combinations.

Full EA-CBM exceeds CEM, ProtoCBM, and PGCM in D-Rate on all six datasets, achieving 78.2--100.0\% versus 51.3--92.6\% for these baselines.
Its margin over the strongest baseline ranges from 3.8 points on FunnyBirds to 21.4 points on Pascal-Part.
It also achieves the highest D-Ratio among all evaluated settings on FunnyBirds, CUB, and CelebA.
Together, Tables~\ref{tab:combinations} and~\ref{tab:response} support complementary roles: matching and grounding improve evidence localization, while matched perturbations strengthen selective responses to region removal.

\section{Results with CLIP RN50}
\label{sec:appendix:rn50}

To assess whether evidence alignment extends beyond ViTs, we repeat Table~\ref{tab:multidataset} with CLIP RN50, whose final stage produces a coarser \(7\times7\) feature grid.

\paragraph{Evaluation Protocol}
EA-CBM trains the final ResNet stage and applies its concept filters to the \(7\times7\) map. 
ProtoCBM uses its released ResNet configuration.
CTS uses Grad-CAM at this stage for all methods; other settings follow ViT-L/14.

\begin{table*}[!t]
\setlength{\aboverulesep}{0pt}\setlength{\belowrulesep}{0pt}
\centering
\small
\setlength{\tabcolsep}{5pt}
\renewcommand{\arraystretch}{1.25}
\caption{\textbf{Classification, concept prediction, and localization with CLIP RN50} (\%). Datasets, splits, baselines, metrics, and the dashes (--) for entries that do not apply follow Table~\ref{tab:multidataset}; EA-CBM and ProtoCBM train the last ResNet stage, and CTS applies Grad-CAM to its \(7\times7\) output. Means over three runs; LP is run once, and the PACO-LVIS values of LP and the label-free, sequential, Minimal-, PCBM-CAV, and SL-CBM baselines come from earlier runs on the same split and protocol. Best values are bold; second-best are underlined.}
\label{tab:rn50}
\vspace{-0.5em}
\resizebox{1.0\textwidth}{!}{%
\begin{tabular}{llcccccccccccccc}
\toprule
\rowcolor{tbhead} &  & \multicolumn{1}{c}{\textbf{Ref.}} & \multicolumn{4}{c}{\textbf{Label-Free}} & \multicolumn{5}{c}{\textbf{Concept-Supervised}} & \multicolumn{3}{c}{\textbf{Spatial}} & \multicolumn{1}{c}{\textbf{Ours}} \\
\headrulefill \cmidrule(lr){3-3} \cmidrule(lr){4-7} \cmidrule(lr){8-12} \cmidrule(lr){13-15} \cmidrule(lr){16-16}
\rowcolor{tbhead} \multirow{-2}{*}{\textbf{Dataset}} & \multirow{-2}{*}{\textbf{Metric}} & \textbf{LP} & \textbf{LF} & \textbf{LM4CV} & \textbf{PS} & \textbf{VLG} & \textbf{Seq.} & \textbf{Joint} & \textbf{CEM} & \textbf{Min.} & \textbf{PCBM} & \textbf{SL} & \textbf{Proto} & \textbf{PGCM} & \textbf{EA-CBM} \\
\midrule
& ACC $\uparrow$ & 73.8 & 71.9 & 72.4 & \secondbest{74.2} & 68.0 & 71.4 & 65.0 & 71.2 & 70.3 & 70.3 & 74.1 & 57.3 & 60.4 & \textbf{76.5}  \\
& C-mAP $\uparrow$ & -- & 10.3 & 6.4 & 47.8 & 38.4 & 52.4 & 74.6 & \secondbest{78.9} & 50.0 & 47.4 & 40.1 & 7.8 & 54.0 & \textbf{81.1}  \\
\multirow{-3}{*}{\textbf{PACO-LVIS}} & CTS $\uparrow$ & -- & 43.2 & 43.1 & 51.9 & 34.4 & 56.5 & \secondbest{57.4} & 56.8 & 48.6 & 56.5 & 51.9 & 41.1 & 46.5 & \textbf{80.4}  \\
\midrule
\rowcolor{tbzebra} & ACC $\uparrow$ & 94.3 & 90.7 & 92.5 & 94.5 & 94.2 & 93.4 & 90.6 & 94.5 & \secondbest{94.7} & 92.9 & 93.0 & 88.6 & 85.4 & \textbf{95.4}  \\
\rowcolor{tbzebra} & C-mAP $\uparrow$ & -- & 41.2 & 38.9 & 74.2 & 81.8 & 84.7 & 85.3 & \secondbest{86.7} & 86.2 & 79.2 & 82.3 & 59.6 & 81.0 & \textbf{90.6}  \\
\rowcolor{tbzebra} \multirow{-3}{*}{\textbf{Pascal-Part}} & CTS $\uparrow$ & -- & 41.5 & 48.0 & 50.6 & \secondbest{69.3} & 56.0 & 57.6 & 60.3 & 62.3 & 54.8 & 47.5 & 52.2 & 56.1 & \textbf{88.2}  \\
\midrule
& ACC $\uparrow$ & 84.7 & 68.8 & 68.3 & 82.9 & 83.1 & 82.3 & 80.4 & 80.8 & \secondbest{88.5} & 82.3 & 78.5 & 84.9 & 79.0 & \textbf{89.6}  \\
& C-mAP $\uparrow$ & -- & 18.9 & 17.5 & 37.4 & 38.4 & 59.2 & 60.9 & 63.5 & 70.3 & 52.5 & 64.2 & 31.1 & \secondbest{70.4} & \textbf{81.2}  \\
\multirow{-3}{*}{\textbf{MonuMAI}} & CTS $\uparrow$ & -- & 37.4 & 36.5 & 44.7 & 46.0 & 68.8 & 69.6 & \secondbest{70.3} & 67.1 & 69.4 & 59.2 & 47.7 & 57.7 & \textbf{88.2}  \\
\midrule
\rowcolor{tbzebra} & ACC $\uparrow$ & 63.8 & 14.4 & 21.5 & 70.9 & 62.0 & 42.8 & 39.7 & 71.3 & 72.7 & 30.3 & 79.1 & \secondbest{84.3} & 71.9 & \textbf{94.9}  \\
\rowcolor{tbzebra} & C-mAP $\uparrow$ & -- & 22.2 & 21.6 & 74.8 & 70.4 & 54.3 & 58.6 & \secondbest{83.4} & 77.9 & 31.3 & 78.8 & 55.0 & 76.8 & \textbf{99.3}  \\
\rowcolor{tbzebra} \multirow{-3}{*}{\textbf{FunnyBirds}} & CTS $\uparrow$ & -- & 47.3 & 37.1 & 73.7 & \secondbest{84.0} & 76.9 & 76.0 & 79.2 & 77.7 & 75.4 & 59.8 & 65.3 & 62.8 & \textbf{98.5}  \\
\midrule
& ACC $\uparrow$ & \secondbest{67.1} & 57.0 & 56.7 & 65.7 & 32.3 & 60.3 & 3.5 & 61.3 & 35.9 & 62.2 & 48.5 & \textbf{68.9} & 47.0 & 54.3  \\
& C-mAP $\uparrow$ & -- & 18.0 & 15.8 & 27.6 & 15.8 & 26.1 & 24.2 & \textbf{29.7} & 27.1 & 25.8 & 25.8 & 16.2 & 24.8 & \secondbest{28.2}  \\
\multirow{-3}{*}{\textbf{CUB}} & CTS $\uparrow$ & -- & 24.2 & 29.7 & 32.0 & 24.8 & 38.2 & 27.8 & 42.0 & 33.7 & 40.9 & 29.2 & 40.7 & \secondbest{53.5} & \textbf{76.2}  \\
\midrule
\rowcolor{tbzebra} & ACC $\uparrow$ & 40.3 & 34.3 & 33.1 & \textbf{45.0} & 36.7 & 37.2 & 36.1 & \secondbest{44.6} & 41.4 & 34.7 & 40.4 & 43.3 & 39.7 & 44.1  \\
\rowcolor{tbzebra} & C-mAP $\uparrow$ & -- & 42.2 & 37.0 & 47.8 & 27.1 & 73.1 & 75.1 & \secondbest{77.2} & 73.6 & 60.9 & 69.5 & 39.9 & 71.5 & \textbf{77.6}  \\
\rowcolor{tbzebra} \multirow{-3}{*}{\textbf{CelebA}} & CTS $\uparrow$ & -- & 90.9 & 78.9 & 79.1 & 17.6 & 88.9 & 91.0 & \secondbest{92.7} & 88.3 & 85.1 & 68.4 & 79.0 & 71.8 & \textbf{99.6}  \\
\midrule
& ACC $\uparrow$ & \textbf{97.5} & 93.9 & 94.4 & \secondbest{97.4} & 95.2 & 94.2 & 4.8 & 94.5 & 94.0 & 94.5 & 91.1 & 95.3 & -- & 96.6  \\
\multirow{-2}{*}{\textbf{Flower102}} & C-mAP $\uparrow$ & -- & -- & -- & -- & -- & 53.6 & 20.0 & \secondbest{95.4} & 58.0 & 81.1 & 66.0 & 7.0 & -- & \textbf{96.8}  \\
\midrule

\rowcolor{tbzebra} & ACC $\uparrow$ & \secondbest{82.4} & 80.6 & 80.8 & \textbf{82.7} & 81.6 & 81.1 & 77.5 & 80.2 & 78.5 & 80.9 & 76.7 & 66.0 & -- & 78.0  \\
\rowcolor{tbzebra} \multirow{-2}{*}{\textbf{Food101}} & C-mAP $\uparrow$ & -- & -- & -- & -- & -- & 63.1 & 74.2 & \textbf{85.0} & 77.6 & 59.7 & 59.2 & 1.9 & -- & \secondbest{81.5}  \\
\bottomrule
\end{tabular}}
\end{table*}

\begin{table*}[!t]
\setlength{\aboverulesep}{0pt}\setlength{\belowrulesep}{0pt}
\centering
\small
\renewcommand{\arraystretch}{1.4}
\caption{\textbf{Accuracy (\%).} Dashes (--) mark entries that do not apply: PGCM needs part regions, which Flower102 and Food101 lack. Mean and sample standard deviation over three runs; LP values without a standard deviation come from one run. Higher is better; best means bold, second-best underlined.}
\label{tab:full:acc}
\vspace{-0.5em}
\resizebox{1.0\textwidth}{!}{%
\begin{tabular}{lcccccccc}
\toprule
\rowcolor{tbhead} \textbf{Method} & \textbf{PACO-LVIS} & \textbf{Pascal-Part} & \textbf{MonuMAI} & \textbf{FunnyBirds} & \textbf{CUB} & \textbf{CelebA} & \textbf{Flower102} & \textbf{Food101} \\
\midrule
\textbf{Linear Probe} & \meanstd{85.47}{0.00} & 95.05 & 88.16 & 78.80 & 84.36 & 39.07 & \secondbest{99.39} & \secondbest{93.05} \\
\midrule
\rowcolor{tbzebra} \textbf{LF-CBM} & \meanstd{81.68}{0.33} & \meanstd{91.40}{0.35} & \meanstd{72.59}{2.72} & \meanstd{19.80}{0.20} & \meanstd{76.08}{0.17} & \meanstd{33.56}{0.29} & \meanstd{97.93}{0.04} & \meanstd{91.70}{0.02} \\
\rowcolor{tbzebra} \textbf{LM4CV} & \meanstd{82.98}{0.19} & \meanstd{91.09}{0.05} & \meanstd{71.86}{0.18} & \meanstd{28.93}{0.23} & \meanstd{73.42}{0.40} & \meanstd{31.41}{0.15} & \meanstd{98.16}{0.37} & \meanstd{91.75}{0.14} \\
\rowcolor{tbzebra} \textbf{PS-CBM} & \meanstd{84.97}{0.13} & \meanstd{94.16}{0.22} & \meanstd{83.90}{1.40} & \meanstd{80.40}{0.53} & \meanstd{82.00}{0.13} & \meanstd{\textbf{42.41}}{0.47} & \meanstd{\secondbest{99.39}}{0.04} & \meanstd{\textbf{93.06}}{0.08} \\
\rowcolor{tbzebra} \textbf{VLG-CBM} & \meanstd{78.07}{0.17} & \meanstd{94.32}{0.14} & \meanstd{87.12}{0.95} & \meanstd{81.00}{2.23} & \meanstd{78.09}{0.64} & \meanstd{35.33}{1.46} & \meanstd{99.38}{0.06} & \meanstd{91.99}{0.36} \\
\midrule
\textbf{Sequential CBM} & \meanstd{82.79}{0.15} & \meanstd{92.96}{0.02} & \meanstd{81.41}{1.30} & \meanstd{64.07}{0.31} & \meanstd{79.74}{0.05} & \meanstd{35.49}{0.11} & \meanstd{98.42}{0.04} & \meanstd{92.43}{0.07} \\
\textbf{Joint CBM} & \meanstd{78.95}{1.49} & \meanstd{89.16}{0.56} & \meanstd{82.97}{3.42} & \meanstd{81.33}{0.64} & \meanstd{57.09}{5.69} & \meanstd{39.06}{0.62} & \meanstd{94.68}{2.07} & \meanstd{93.00}{0.03} \\
\textbf{CEM} & \meanstd{81.65}{0.16} & \meanstd{94.63}{0.07} & \meanstd{86.19}{1.83} & \meanstd{84.47}{1.33} & \meanstd{79.57}{0.75} & \meanstd{38.47}{1.18} & \meanstd{98.52}{0.16} & \meanstd{91.18}{0.39} \\
\textbf{Minimal-CBM} & \meanstd{83.03}{0.22} & \meanstd{94.70}{0.16} & \meanstd{86.40}{0.95} & \meanstd{85.53}{0.12} & \meanstd{61.44}{1.35} & \meanstd{41.45}{0.29} & \meanstd{98.70}{0.15} & \meanstd{91.51}{0.23} \\
\textbf{PCBM-CAV} & \meanstd{83.65}{0.21} & \meanstd{93.04}{0.20} & \meanstd{82.04}{1.26} & \meanstd{58.87}{1.14} & \meanstd{82.49}{0.10} & \meanstd{34.02}{0.22} & \meanstd{98.94}{0.04} & \meanstd{92.20}{0.05} \\
\midrule
\rowcolor{tbzebra} \textbf{SL-CBM} & \meanstd{87.92}{0.43} & \meanstd{\secondbest{95.30}}{0.10} & \meanstd{84.94}{1.09} & \meanstd{81.40}{4.39} & \meanstd{68.04}{1.73} & \meanstd{39.78}{0.74} & \meanstd{98.85}{0.41} & \meanstd{90.36}{0.27} \\
\rowcolor{tbzebra} \textbf{ProtoCBM} & \meanstd{\secondbest{88.00}}{0.41} & \meanstd{\textbf{96.54}}{0.07} & \meanstd{\textbf{92.83}}{0.62} & \meanstd{\textbf{94.87}}{0.76} & \meanstd{\secondbest{85.60}}{0.23} & \meanstd{\secondbest{42.17}}{0.27} & \meanstd{\textbf{99.49}}{0.08} & \meanstd{92.70}{0.04} \\
\rowcolor{tbzebra} \textbf{PGCM} & \meanstd{77.19}{0.84} & \meanstd{87.58}{0.28} & \meanstd{88.47}{1.73} & \meanstd{81.33}{1.90} & \meanstd{65.99}{0.56} & \meanstd{38.61}{0.68} & -- & -- \\
\midrule

\textbf{EA-CBM} & \meanstd{\textbf{88.36}}{0.23} & \meanstd{\textbf{96.54}}{0.04} & \meanstd{\secondbest{92.11}}{1.26} & \meanstd{\secondbest{94.67}}{0.12} & \meanstd{\textbf{85.97}}{0.04} & \meanstd{40.87}{0.43} & \meanstd{99.26}{0.05} & \meanstd{92.50}{0.02}  \\
\bottomrule
\end{tabular}}
\end{table*}

\begin{table*}[!t]
\setlength{\aboverulesep}{0pt}\setlength{\belowrulesep}{0pt}
\centering
\small
\renewcommand{\arraystretch}{1.5}
\caption{\textbf{Macro-F1 (\%).} Dashes (--) mark entries that do not apply: PGCM needs part regions, which Flower102 and Food101 lack. Mean and sample standard deviation over three runs; LP values without a standard deviation come from one run. Higher is better; best means bold, second-best underlined.}
\label{tab:full:f1}
\vspace{-0.5em}
\resizebox{1.0\textwidth}{!}{%
\begin{tabular}{lcccccccc}
\toprule
\rowcolor{tbhead} \textbf{Method} & \textbf{PACO-LVIS} & \textbf{Pascal-Part} & \textbf{MonuMAI} & \textbf{FunnyBirds} & \textbf{CUB} & \textbf{CelebA} & \textbf{Flower102} & \textbf{Food101} \\
\midrule
\textbf{Linear Probe} & \meanstd{82.51}{0.00} & 95.04 & 88.34 & 78.46 & 84.47 & 10.88 & \textbf{99.43} & \secondbest{93.07} \\
\midrule
\rowcolor{tbzebra} \textbf{LF-CBM} & \meanstd{79.16}{0.32} & \meanstd{90.88}{0.40} & \meanstd{70.20}{2.78} & \meanstd{18.37}{0.11} & \meanstd{76.24}{0.18} & \meanstd{7.91}{0.44} & \meanstd{97.69}{0.07} & \meanstd{91.72}{0.02} \\
\rowcolor{tbzebra} \textbf{LM4CV} & \meanstd{80.62}{0.23} & \meanstd{90.29}{0.07} & \meanstd{67.86}{1.39} & \meanstd{27.57}{0.42} & \meanstd{73.45}{0.36} & \meanstd{8.13}{0.10} & \meanstd{98.01}{0.41} & \meanstd{91.74}{0.14} \\
\rowcolor{tbzebra} \textbf{PS-CBM} & \meanstd{82.51}{0.29} & \meanstd{93.82}{0.27} & \meanstd{84.53}{1.22} & \meanstd{79.96}{0.68} & \meanstd{82.18}{0.16} & \meanstd{16.69}{1.74} & \meanstd{99.34}{0.03} & \meanstd{\textbf{93.09}}{0.08} \\
\rowcolor{tbzebra} \textbf{VLG-CBM} & \meanstd{75.53}{0.35} & \meanstd{94.09}{0.11} & \meanstd{87.86}{1.05} & \meanstd{80.96}{2.23} & \meanstd{78.09}{0.59} & \meanstd{8.72}{0.64} & \meanstd{99.37}{0.09} & \meanstd{92.02}{0.36} \\
\midrule
\textbf{Sequential CBM} & \meanstd{80.23}{0.33} & \meanstd{91.44}{0.09} & \meanstd{80.95}{1.18} & \meanstd{63.86}{0.35} & \meanstd{79.76}{0.04} & \meanstd{9.65}{0.12} & \meanstd{98.29}{0.08} & \meanstd{92.44}{0.07} \\
\textbf{Joint CBM} & \meanstd{68.41}{4.31} & \meanstd{78.52}{3.41} & \meanstd{81.23}{5.52} & \meanstd{81.03}{0.64} & \meanstd{53.31}{6.09} & \meanstd{7.85}{0.65} & \meanstd{91.04}{3.60} & \meanstd{93.03}{0.03} \\
\textbf{CEM} & \meanstd{79.70}{0.93} & \meanstd{94.54}{0.04} & \meanstd{86.13}{2.28} & \meanstd{84.21}{1.46} & \meanstd{79.05}{0.97} & \meanstd{15.27}{2.11} & \meanstd{98.40}{0.19} & \meanstd{91.21}{0.38} \\
\textbf{Minimal-CBM} & \meanstd{80.18}{0.57} & \meanstd{94.61}{0.16} & \meanstd{85.81}{1.64} & \meanstd{85.51}{0.24} & \meanstd{60.83}{1.22} & \meanstd{14.96}{0.25} & \meanstd{98.53}{0.24} & \meanstd{91.53}{0.23} \\
\textbf{PCBM-CAV} & \meanstd{81.29}{0.58} & \meanstd{92.49}{0.14} & \meanstd{81.27}{2.74} & \meanstd{58.20}{0.81} & \meanstd{82.50}{0.06} & \meanstd{9.64}{0.66} & \meanstd{98.97}{0.07} & \meanstd{92.20}{0.04} \\
\midrule
\rowcolor{tbzebra} \textbf{SL-CBM} & \meanstd{\secondbest{85.85}}{0.20} & \meanstd{95.09}{0.09} & \meanstd{83.77}{1.72} & \meanstd{81.19}{4.47} & \meanstd{68.06}{1.78} & \meanstd{12.77}{0.06} & \meanstd{98.81}{0.26} & \meanstd{90.37}{0.27} \\
\rowcolor{tbzebra} \textbf{ProtoCBM} & \meanstd{85.65}{0.44} & \meanstd{\textbf{96.56}}{0.13} & \meanstd{\textbf{93.06}}{0.72} & \meanstd{\textbf{94.82}}{0.77} & \meanstd{\secondbest{85.51}}{0.28} & \meanstd{\textbf{20.80}}{0.78} & \meanstd{\secondbest{99.42}}{0.08} & \meanstd{92.70}{0.04} \\
\rowcolor{tbzebra} \textbf{PGCM} & \meanstd{73.64}{0.99} & \meanstd{80.12}{1.45} & \meanstd{88.60}{1.66} & \meanstd{81.00}{2.08} & \meanstd{65.90}{0.39} & \meanstd{15.21}{1.23} & -- & -- \\
\midrule

\textbf{EA-CBM} & \meanstd{\textbf{86.48}}{0.40} & \meanstd{\secondbest{96.53}}{0.02} & \meanstd{\secondbest{92.63}}{1.14} & \meanstd{\secondbest{94.64}}{0.13} & \meanstd{\textbf{86.01}}{0.03} & \meanstd{\secondbest{18.21}}{0.08} & \meanstd{99.19}{0.01} & \meanstd{92.52}{0.02}  \\
\bottomrule
\end{tabular}}
\end{table*}

\begin{table*}[!t]
\setlength{\aboverulesep}{0pt}\setlength{\belowrulesep}{0pt}
\centering
\small
\renewcommand{\arraystretch}{1.5}
\caption{\textbf{Concept AUROC (\%).} The linear probe (no concepts) is omitted. Dashes (--) mark inapplicable entries: on Flower102 and Food101, label-free CBMs replace the class-level concept lists with their own, and PGCM needs part regions. Mean and sample standard deviation over three runs. Higher is better; best means bold, second-best underlined.}
\label{tab:full:auroc}
\vspace{-0.5em}
\resizebox{1.0\textwidth}{!}{%
\begin{tabular}{lcccccccc}
\toprule
\rowcolor{tbhead} \textbf{Method} & \textbf{PACO-LVIS} & \textbf{Pascal-Part} & \textbf{MonuMAI} & \textbf{FunnyBirds} & \textbf{CUB} & \textbf{CelebA} & \textbf{Flower102} & \textbf{Food101} \\
\midrule
\textbf{LF-CBM} & \meanstd{68.3}{0.1} & \meanstd{70.3}{0.3} & \meanstd{60.5}{0.8} & \meanstd{55.1}{0.1} & \meanstd{63.7}{\textless0.1} & \meanstd{66.6}{0.3} & -- & -- \\
\textbf{LM4CV} & \meanstd{66.9}{\textless0.1} & \meanstd{68.9}{\textless0.1} & \meanstd{57.1}{\textless0.1} & \meanstd{54.3}{\textless0.1} & \meanstd{64.7}{\textless0.1} & \meanstd{64.7}{\textless0.1} & -- & -- \\
\textbf{PS-CBM} & \meanstd{94.2}{6.8} & \meanstd{92.5}{\textless0.1} & \meanstd{76.7}{0.1} & \meanstd{95.7}{0.7} & \meanstd{72.8}{0.1} & \meanstd{77.0}{0.1} & -- & -- \\
\textbf{VLG-CBM} & \meanstd{85.1}{8.8} & \meanstd{95.9}{0.1} & \meanstd{78.2}{0.1} & \meanstd{97.3}{0.4} & \meanstd{63.2}{0.1} & \meanstd{57.2}{\textless0.1} & -- & -- \\
\midrule
\rowcolor{tbzebra} \textbf{Sequential CBM} & \meanstd{97.8}{\textless0.1} & \meanstd{95.6}{\textless0.1} & \meanstd{88.5}{\textless0.1} & \meanstd{89.1}{\textless0.1} & \meanstd{76.1}{\textless0.1} & \meanstd{90.2}{\textless0.1} & \meanstd{99.3}{\textless0.1} & \meanstd{99.5}{\textless0.1} \\
\rowcolor{tbzebra} \textbf{Joint CBM} & \meanstd{76.4}{0.5} & \meanstd{88.6}{0.6} & \meanstd{77.3}{0.7} & \meanstd{87.9}{1.9} & \meanstd{74.0}{0.2} & \meanstd{89.1}{0.1} & \meanstd{89.2}{0.8} & \meanstd{\textbf{99.8}}{\textless0.1} \\
\rowcolor{tbzebra} \textbf{CEM} & \meanstd{\secondbest{99.3}}{\textless0.1} & \meanstd{96.2}{0.4} & \meanstd{91.1}{1.2} & \meanstd{\secondbest{99.0}}{0.1} & \meanstd{\secondbest{77.0}}{0.2} & \meanstd{\textbf{91.8}}{0.2} & \meanstd{\textbf{100.0}}{\textless0.1} & \meanstd{\textbf{99.8}}{\textless0.1} \\
\rowcolor{tbzebra} \textbf{Minimal-CBM} & \meanstd{97.8}{0.3} & \meanstd{\secondbest{96.3}}{0.5} & \meanstd{92.1}{0.3} & \meanstd{98.6}{0.1} & \meanstd{74.9}{0.2} & \meanstd{\secondbest{91.2}}{0.7} & \meanstd{92.0}{0.4} & \meanstd{98.2}{0.2} \\
\rowcolor{tbzebra} \textbf{PCBM-CAV} & \meanstd{97.7}{\textless0.1} & \meanstd{94.4}{\textless0.1} & \meanstd{89.1}{0.9} & \meanstd{85.3}{0.1} & \meanstd{72.7}{0.1} & \meanstd{84.8}{0.2} & \meanstd{\secondbest{99.9}}{\textless0.1} & \meanstd{99.6}{\textless0.1} \\
\midrule
\textbf{SL-CBM} & \meanstd{96.3}{2.9} & \meanstd{95.0}{0.3} & \meanstd{90.6}{0.3} & \meanstd{90.5}{0.9} & \meanstd{70.2}{0.3} & \meanstd{87.7}{0.3} & \meanstd{99.8}{0.2} & \meanstd{98.5}{\textless0.1} \\
\textbf{ProtoCBM} & \meanstd{87.7}{2.1} & \meanstd{95.9}{0.2} & \meanstd{\textbf{95.6}}{0.9} & \meanstd{89.4}{1.9} & \meanstd{71.2}{0.6} & \meanstd{75.2}{0.8} & \meanstd{94.8}{0.4} & \meanstd{64.3}{0.4} \\
\textbf{PGCM} & \meanstd{97.0}{0.2} & \meanstd{95.4}{0.4} & \meanstd{90.7}{1.4} & \meanstd{90.0}{0.5} & \meanstd{73.1}{1.1} & \meanstd{90.9}{0.2} & -- & -- \\
\midrule

\rowcolor{tbzebra} \textbf{EA-CBM} & \meanstd{\textbf{99.7}}{\textless0.1} & \meanstd{\textbf{97.8}}{\textless0.1} & \meanstd{\secondbest{94.2}}{0.8} & \meanstd{\textbf{99.8}}{\textless0.1} & \meanstd{\textbf{78.4}}{0.1} & \meanstd{90.7}{0.2} & \meanstd{\textbf{100.0}}{\textless0.1} & \meanstd{\secondbest{99.7}}{\textless0.1}  \\
\bottomrule
\end{tabular}}
\end{table*}

\begin{table*}[!t]
\setlength{\aboverulesep}{0pt}\setlength{\belowrulesep}{0pt}
\centering
\small
\renewcommand{\arraystretch}{1.5}
\caption{\textbf{Concept mAP (\%).} The linear probe (no concepts) is omitted. Dashes (--) mark inapplicable entries: on Flower102 and Food101, label-free CBMs replace the class-level concept lists with their own, and PGCM needs part regions. Mean and sample standard deviation over three runs. Higher is better; best means bold, second-best underlined.}
\label{tab:full:map}
\vspace{-0.5em}
\resizebox{1.0\textwidth}{!}{%
\begin{tabular}{lcccccccc}
\toprule
\rowcolor{tbhead} \textbf{Method} & \textbf{PACO-LVIS} & \textbf{Pascal-Part} & \textbf{MonuMAI} & \textbf{FunnyBirds} & \textbf{CUB} & \textbf{CelebA} & \textbf{Flower102} & \textbf{Food101} \\
\midrule
\textbf{LF-CBM} & \meanstd{10.7}{0.1} & \meanstd{41.5}{0.1} & \meanstd{21.3}{0.1} & \meanstd{22.9}{0.1} & \meanstd{18.2}{0.1} & \meanstd{40.0}{0.3} & -- & -- \\
\textbf{LM4CV} & \meanstd{10.5}{\textless0.1} & \meanstd{38.1}{\textless0.1} & \meanstd{20.8}{\textless0.1} & \meanstd{22.4}{\textless0.1} & \meanstd{18.7}{\textless0.1} & \meanstd{36.5}{\textless0.1} & -- & -- \\
\textbf{PS-CBM} & \meanstd{49.8}{6.5} & \meanstd{72.9}{0.2} & \meanstd{37.2}{0.6} & \meanstd{82.0}{2.0} & \meanstd{27.8}{\textless0.1} & \meanstd{45.9}{0.2} & -- & -- \\
\textbf{VLG-CBM} & \meanstd{37.0}{8.5} & \meanstd{81.0}{0.2} & \meanstd{39.9}{0.1} & \meanstd{89.0}{1.5} & \meanstd{19.5}{0.2} & \meanstd{27.4}{\textless0.1} & -- & -- \\
\midrule
\rowcolor{tbzebra} \textbf{Sequential CBM} & \meanstd{57.9}{\textless0.1} & \meanstd{84.0}{\textless0.1} & \meanstd{56.9}{\textless0.1} & \meanstd{66.6}{\textless0.1} & \meanstd{28.2}{\textless0.1} & \meanstd{69.9}{\textless0.1} & \meanstd{79.3}{\textless0.1} & \meanstd{89.6}{\textless0.1} \\
\rowcolor{tbzebra} \textbf{Joint CBM} & \meanstd{46.2}{1.3} & \meanstd{72.3}{2.4} & \meanstd{50.7}{0.5} & \meanstd{73.1}{3.1} & \meanstd{\secondbest{31.6}}{\textless0.1} & \meanstd{69.1}{0.9} & \meanstd{68.1}{0.9} & \meanstd{\textbf{96.1}}{\textless0.1} \\
\rowcolor{tbzebra} \textbf{CEM} & \meanstd{\secondbest{86.8}}{0.2} & \meanstd{\secondbest{86.1}}{1.1} & \meanstd{67.7}{6.5} & \meanstd{\secondbest{95.4}}{0.3} & \meanstd{30.2}{0.1} & \meanstd{\secondbest{72.9}}{0.3} & \meanstd{\secondbest{99.4}}{0.1} & \meanstd{\secondbest{95.8}}{\textless0.1} \\
\rowcolor{tbzebra} \textbf{Minimal-CBM} & \meanstd{54.3}{0.5} & \meanstd{85.6}{0.9} & \meanstd{71.0}{2.4} & \meanstd{93.4}{0.3} & \meanstd{28.8}{1.0} & \meanstd{71.1}{1.4} & \meanstd{76.2}{1.1} & \meanstd{91.6}{0.6} \\
\rowcolor{tbzebra} \textbf{PCBM-CAV} & \meanstd{54.2}{0.6} & \meanstd{78.7}{0.3} & \meanstd{63.1}{1.1} & \meanstd{57.4}{1.0} & \meanstd{23.1}{0.1} & \meanstd{57.9}{0.1} & \meanstd{98.0}{0.1} & \meanstd{88.8}{\textless0.1} \\
\midrule
\textbf{SL-CBM} & \meanstd{54.8}{3.0} & \meanstd{82.8}{0.1} & \meanstd{70.0}{0.4} & \meanstd{80.1}{3.3} & \meanstd{26.5}{0.6} & \meanstd{68.5}{0.6} & \meanstd{94.6}{2.7} & \meanstd{90.0}{0.6} \\
\textbf{ProtoCBM} & \meanstd{49.3}{1.9} & \meanstd{81.6}{0.8} & \meanstd{\secondbest{79.3}}{5.6} & \meanstd{72.3}{2.8} & \meanstd{25.9}{0.6} & \meanstd{42.1}{0.5} & \meanstd{55.7}{1.4} & \meanstd{4.5}{0.2} \\
\textbf{PGCM} & \meanstd{75.2}{1.8} & \meanstd{85.7}{0.4} & \meanstd{78.9}{1.0} & \meanstd{82.1}{0.4} & \meanstd{27.1}{0.3} & \meanstd{69.2}{0.6} & -- & -- \\
\midrule

\rowcolor{tbzebra} \textbf{EA-CBM} & \meanstd{\textbf{88.4}}{0.1} & \meanstd{\textbf{90.2}}{0.2} & \meanstd{\textbf{80.8}}{1.6} & \meanstd{\textbf{99.3}}{0.2} & \meanstd{\textbf{32.1}}{0.1} & \meanstd{\textbf{73.2}}{0.3} & \meanstd{\textbf{99.5}}{\textless0.1} & \meanstd{95.6}{\textless0.1}  \\
\bottomrule
\end{tabular}}
\end{table*}

\begin{table*}[!t]
\setlength{\aboverulesep}{0pt}\setlength{\belowrulesep}{0pt}
\centering
\small
\renewcommand{\arraystretch}{1.4}
\caption{\textbf{CTS (\%).} The linear probe (no concepts) and Flower102 and Food101 (no regions) are omitted. Mean and sample standard deviation over three runs. CTS uses Grad-CAM at the features each concept head reads, and at the final-block input for SL-CBM on PACO-LVIS. Higher is better; best means bold, second-best underlined.}
\label{tab:full:cts}
\vspace{-0.5em}
\begin{tabular}{lcccccc}
\toprule
\rowcolor{tbhead} \textbf{Method} & \textbf{PACO-LVIS} & \textbf{Pascal-Part} & \textbf{MonuMAI} & \textbf{FunnyBirds} & \textbf{CUB} & \textbf{CelebA} \\
\midrule
\textbf{LF-CBM} & \meanstd{55.5}{0.8} & \meanstd{57.6}{0.9} & \meanstd{63.8}{1.6} & \meanstd{66.5}{0.1} & \meanstd{34.3}{0.4} & \meanstd{\secondbest{91.3}}{1.6} \\
\textbf{LM4CV} & \meanstd{\secondbest{56.0}}{\textless0.1} & \meanstd{58.6}{\textless0.1} & \meanstd{51.3}{\textless0.1} & \meanstd{72.8}{\textless0.1} & \meanstd{32.0}{\textless0.1} & \meanstd{83.2}{\textless0.1} \\
\textbf{PS-CBM} & \meanstd{47.9}{5.7} & \meanstd{53.8}{0.8} & \meanstd{47.6}{0.5} & \meanstd{61.7}{1.4} & \meanstd{34.3}{0.5} & \meanstd{78.7}{0.9} \\
\textbf{VLG-CBM} & \meanstd{30.3}{0.8} & \meanstd{56.5}{0.8} & \meanstd{44.7}{1.0} & \meanstd{61.8}{2.3} & \meanstd{22.4}{0.6} & \meanstd{17.5}{0.1} \\
\midrule
\rowcolor{tbzebra} \textbf{Sequential CBM} & \meanstd{52.0}{\textless0.1} & \meanstd{55.7}{0.1} & \meanstd{\secondbest{64.5}}{0.3} & \meanstd{\secondbest{88.2}}{\textless0.1} & \meanstd{\secondbest{56.2}}{\textless0.1} & \meanstd{85.9}{\textless0.1} \\
\rowcolor{tbzebra} \textbf{Joint CBM} & \meanstd{45.1}{0.9} & \meanstd{\secondbest{59.5}}{0.4} & \meanstd{49.5}{1.3} & \meanstd{69.7}{2.3} & \meanstd{53.1}{0.7} & \meanstd{86.4}{1.4} \\
\rowcolor{tbzebra} \textbf{CEM} & \meanstd{52.7}{0.1} & \meanstd{56.2}{3.3} & \meanstd{62.4}{2.7} & \meanstd{77.5}{0.7} & \meanstd{47.2}{1.3} & \meanstd{86.8}{0.3} \\
\rowcolor{tbzebra} \textbf{Minimal-CBM} & \meanstd{46.8}{1.4} & \meanstd{59.3}{1.7} & \meanstd{58.5}{2.1} & \meanstd{75.4}{1.5} & \meanstd{36.0}{3.0} & \meanstd{85.8}{0.7} \\
\rowcolor{tbzebra} \textbf{PCBM-CAV} & \meanstd{51.7}{0.7} & \meanstd{51.7}{2.5} & \meanstd{60.3}{1.1} & \meanstd{75.6}{1.5} & \meanstd{36.5}{0.1} & \meanstd{79.4}{0.8} \\
\midrule
\textbf{SL-CBM} & \meanstd{43.4}{6.6} & \meanstd{42.9}{1.3} & \meanstd{52.4}{1.2} & \meanstd{47.7}{7.2} & \meanstd{24.3}{0.4} & \meanstd{66.7}{1.1} \\
\textbf{ProtoCBM} & \meanstd{32.7}{0.8} & \meanstd{36.0}{5.7} & \meanstd{61.8}{0.8} & \meanstd{55.6}{6.5} & \meanstd{27.6}{2.4} & \meanstd{62.9}{3.8} \\
\textbf{PGCM} & \meanstd{46.2}{0.8} & \meanstd{50.9}{0.3} & \meanstd{62.2}{3.0} & \meanstd{60.4}{3.0} & \meanstd{51.7}{4.6} & \meanstd{73.0}{1.8} \\
\midrule

\rowcolor{tbzebra} \textbf{EA-CBM} & \meanstd{\textbf{83.2}}{0.8} & \meanstd{\textbf{89.7}}{\textless0.1} & \meanstd{\textbf{91.0}}{0.2} & \meanstd{\textbf{99.4}}{0.2} & \meanstd{\textbf{98.4}}{\textless0.1} & \meanstd{\textbf{99.9}}{\textless0.1} \\
\bottomrule
\end{tabular}
\end{table*}

\paragraph{CLIP RN50 Results}
In Table~\ref{tab:rn50}, EA-CBM leads CTS on all six datasets with regions, from 76.2 on CUB to 99.6 on CelebA.
Its gains over the strongest baseline are 6.9--23.0 points, largest on PACO-LVIS, CUB, and Pascal-Part, as with ViT-L/14.
It also leads C-mAP on six of eight datasets and ACC on PACO-LVIS, Pascal-Part, MonuMAI, and FunnyBirds, supporting evidence alignment with convolutional features.
On Flower102, it has the highest C-mAP (96.8).

\section{Complete Results}
\label{sec:appendix:full}

Tables~\ref{tab:full:acc}--\ref{tab:full:cts} extend Table~\ref{tab:multidataset} with macro-F1 and concept AUROC, complementary measures of predictive performance and variability across runs.
We report means and sample standard deviations over three runs; the linear probe uses one run.

EA-CBM achieves the highest ACC and macro-F1 on PACO-LVIS and CUB, ties ProtoCBM in ACC on Pascal-Part, and ranks second in both metrics on MonuMAI and FunnyBirds (Tables~\ref{tab:full:acc} and~\ref{tab:full:f1}).
The macro-F1 results complement ACC by weighting classes equally, showing that EA-CBM remains competitive under both evaluation criteria.
For concept prediction, EA-CBM achieves the highest C-mAP on seven of eight datasets and the highest reported concept AUROC on five, including a tie on Flower102 (Tables~\ref{tab:full:auroc} and~\ref{tab:full:map}).
These metrics support the predictive quality of its concept responses.

EA-CBM also achieves the highest CTS on all six datasets with regions, combining strong mean localization with low variation across runs (Table~\ref{tab:full:cts}).
Its CTS standard deviations are at most 0.8 points and below 0.1 on Pascal-Part, CUB, and CelebA.
Baseline CTS is more variable in some settings, reaching a standard deviation of 7.2 points for SL-CBM on FunnyBirds.
The localization margins in Section~\ref{sec:expt:localization} exceed EA-CBM's largest CTS standard deviation by more than tenfold, indicating that its gains are substantial relative to its variation across seeds.

\end{document}